%% file: main.tex
\documentclass[10pt,twocolumn,letterpaper]{article}

\usepackage{cvpr}      % REVIEW version
\input{preamble}
\usepackage{iftex}
\ifXeTeX
  \usepackage{fontspec}
\fi
\usepackage{booktabs}
\usepackage{multirow}
\usepackage{graphicx}
\usepackage{amsmath,amssymb}
\usepackage{xspace}
\usepackage{enumitem}
\usepackage{float}

\newcommand{\ours}{Tracer\xspace}
\newcommand{\bench}{Retrace\xspace}
\graphicspath{{fig/}}

\definecolor{cvprblue}{rgb}{0.21,0.49,0.74}
\usepackage[pagebackref,breaklinks,colorlinks,allcolors=cvprblue]{hyperref}

\title{Where Have I Been? Benchmarking Trajectory-Level Spatial Reasoning\\in Egocentric Walkthroughs}

\author{
  {\bf Zizhong Li}$^\ast$, {\bf Haopeng Zhang}$^\dagger$, {\bf Jiawei Zhang}$^\ast$ \\
  $^\ast$IFM Lab, University of California, Davis \\
  $^\dagger$School of Data Science, University of North Carolina at Charlotte \\
  \texttt{zzoli@ucdavis.edu, haopeng.zhang@charlotte.edu, jiawei@ifmlab.org}
}

\begin{document}
\maketitle
\input{sec/0_teaser}
\input{sec/0_abstract}

\input{sec/1_intro}

\input{sec/2_related}
\input{sec/3_benchmark}
\input{sec/4_method}
\input{sec/5_experiment_v2}
\input{sec/6_conclusion}
{
    \small
    \bibliographystyle{ieeenat_fullname}
    \bibliography{main}
}

% Supplementary material (split the PDF after the references for submission).
\input{sec/X_suppl}
\end{document}

%% file: preamble.tex
\usepackage{cuted} %< for strip environment

\usepackage{currfile} %< for \currfilebase macro

\usepackage{caption} %< for "\captionof" commands (teaser)

%% file: sec/0_teaser.tex
% Full-width teaser (fig/teaser_v27.pdf, make_teaser_v27_fig.py): (a) and (c) are Tracer outputs on the test split.
% Retrace evaluates not only what a model answers about an egocentric walkthrough, but also which places and visits support that answer.
\begin{strip}
\centering
\includegraphics[width=1.0\textwidth]{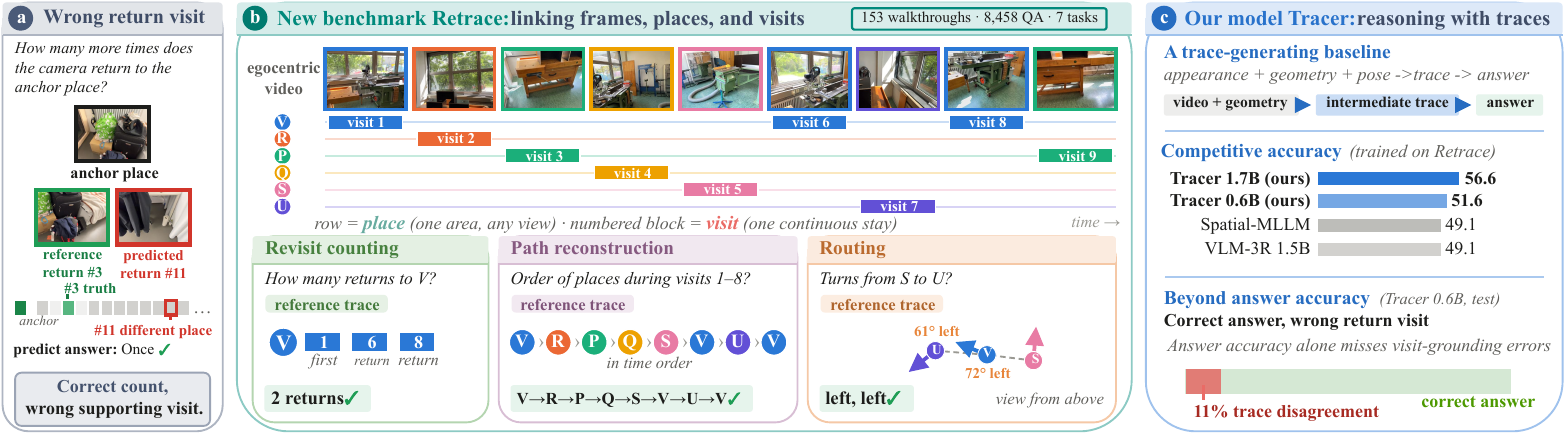}
\captionof{figure}{\textbf{A correct answer does not guarantee correct place recognition or visit tracking.}
(a)~The model correctly answers ``Once,'' but grounds the return at a different place.
(b)~Our benchmark \textbf{\bench} links video frames to \emph{places} and distinct \emph{visits}, enabling checks of spatial predictions beyond answer accuracy.
(c)~Our model \textbf{\ours} generates intermediate traces before answering and achieves competitive accuracy, yet there is still 11\% ground returns differently from \bench's reference visits among its correct revisit answers.}
\label{fig:teaser}
% \vspace{-1mm}
\end{strip}

%% file: sec/0_abstract.tex
\begin{abstract}
    Understanding an egocentric walkthrough requires recognizing places across viewpoints, distinguishing repeated visits, and tracking the route between them. 
    However, a correct final answer alone cannot establish whether a model correctly identifies these intermediate facts. 
    To evaluate this \emph{trajectory-level spatial reasoning}, we introduce \textbf{Retrace}, a benchmark comprising 8,458 multiple-choice questions across seven task types over 153 ScanNet++ walkthroughs. 
    Its core is a shared reference trace that links video frames to human-reviewed places, distinct visits in temporal order, and observed transitions.
    This reference supports the evaluation of explicit spatial predictions alongside final answers alone, including place reconstruction, routing, and the observations identified as return visits.
    We also introduce \textbf{\ours}, a trace-generating baseline that combines appearance, geometry, and camera pose features, using reference trace supervision to predict task-relevant spatial facts before answering.
    On the scene-disjoint test set, \ours reaches 56.6\% answer accuracy with a 1.7B language model backbone, compared with 49.9\% for the best of nine zero-shot general-purpose MLLMs and 49.1\% for the best of 3D-aware baselines.
    Beyond answer accuracy, further analysis on \bench reveals that correct answers can accompany intermediate reasoning traces with incorrect place or turn sequences, as observed in both \ours and other baselines.
    % \bench further exposes incorrect return grounding behind correct
    % revisit counts that count-only options cannot reveal.
    Together, these findings demonstrate the value of evaluating sequence recovery and visit grounding alongside answer accuracy.
\end{abstract}

%% file: sec/1_intro.tex
\section{Introduction}
\label{sec:intro}
After walking through an unfamiliar building, you may recognize a corridor when you return to it.
But when did you first reach it, and which turns did you take before returning?
In an egocentric walkthrough, the same place may appear at different times and from different viewpoints.
Answering such questions requires recognizing the same place across views while distinguishing separate visits and tracking the route between them.
We call this capability \emph{trajectory-level spatial reasoning} (Fig.~\ref{fig:teaser}).

Multimodal large language models (MLLMs) have made rapid progress in visual and video understanding~\cite{liu2023llava, zhang2024llava, bai2025qwen25vl}.
Existing benchmarks evaluate spatial measurements, order of appearance, and route planning~\cite{yang2025thinking}, while recent work also covers revisits and chronological spatial recall~\cite{li2026ovo}.
Some further provide human-written reasoning annotations~\cite{yang2026mmsibench} or temporal evidence intervals~\cite{li2026ovo}.
However, \textit{a correct final answer alone does not establish whether a model correctly matched views of the same place, distinguished separate visits, or recovered the relevant route}.
For example, a model may count returns correctly while grounding them at wrong places. 
Evaluating such intermediate spatial predictions requires a reference linking video frames to
places, distinct visits, and observed transitions, so that predicted visits and routes can be
checked alongside final answers (Tab.~\ref{tab:bench_compare}).
% We therefore seek to evaluate explicit spatial predictions against a shared reference linking video frames to places, distinct visits, and observed transitions.
% A correct path option specifies a place sequence, but does not identify which observations support each visit.
% Likewise, a correct revisit count does not establish whether the predicted returns correspond to the queried place. 
% \ours links these predictions to frames and distinct visits, enabling evaluation of their spatial grounding alongside answer correctness.
% A shared reference linking places, visits, and observed transitions to video frames enables consistent evaluation of these intermediate predictions across question types (Tab.~\ref{tab:bench_compare}).
% We therefore seek to evaluate not only the final answer, but also which frames a model associates with the same place, how it distinguishes separate visits, and which transitions it predicts between them.
% This requires a shared reference linking places, visits, and observed transitions to video frames (Tab.~\ref{tab:bench_compare}).

To support this evaluation, we introduce \textbf{\bench}, a benchmark comprising 8{,}458 multiple-choice questions over 153 ScanNet++~\cite{yeshwanth2023scannetpp} egocentric walkthroughs.
Its reference distinguishes a \emph{place}, a local area that may appear across multiple views, from a \emph{visit}, a consecutive sequence of observations of that place.
A later return to the same place constitutes a separate visit.
For each walkthrough, we record places, ordered visits, and observed transitions, together with metric geometry from depth and camera poses.
This shared reference supports seven question types spanning spatial relations, visit history, and routes.
Each question includes a final answer and a task-specific trace specifying the relevant visits, place sequences, turns, or spatial relations, linked to video frames where applicable.
These reference traces can support both intermediate evaluation and training supervision, and are not provided at test time.
% This representation supports five trajectory-level tasks: temporal ordering, temporal localization, revisit counting, routing, and path reconstruction, along with two pairwise tasks, adjacency and relative distance.
% Each question in \bench is paired with a graph-derived answer and a frame-linked trace of the relevant visits.
% These traces provide intermediate supervision for training and reference targets for evaluating intermediate predictions. 
% The place graph and visit annotations are not provided as inputs at evaluation.

Recent 3D-aware MLLMs incorporate geometric information through pretrained encoders~\cite{wu2025spatialmllm, fan2026vlm} or 3D positional encodings~\cite{zheng2025video3dllm}.
% We use explicit geometric objectives and the frame-linked traces in \bench to address two requirements of trajectory-level reasoning: relating observations across viewpoints and tracking separate visits along a walk.
To study how geometry and reference-derived traces support trajectory-level spatial reasoning, we introduce \textbf{\ours}, a trace-generating multimodal baseline built on small language models.
It combines appearance with depth-derived geometry and camera poses expressed in a common scene coordinate.
The training for \ours proceeds in two stages.
First, geometric pretraining trains \ours to recover masked 3D coordinates, match corresponding patches across frames, and estimate relative camera poses.
Task-specific fine-tuning then trains the model to generate frame-referenced intermediate reasoning traces before selecting an answer.

We evaluate general-purpose and 3D-aware MLLMs on \bench.
After the two-stage training, \ours achieves 51.6\% overall answer accuracy with a 0.6B language model and 56.6\% with a 1.7B model, compared with 49.9\% for the best of nine zero-shot general-purpose MLLMs and 49.1\% for the best finetuned 3D-aware  baselines.
% MLLM reveals a gap between selecting correct options and
% independently reconstructing place and turn sequences.
Beyond answer accuracy, the analysis of \ours reveals that correct answers can accompany intermediate reasoning traces with incorrect place or turn sequences, and the frame-linked trace reference in \bench further enables more precise verification of where the predicted returns occur, even when the revisit count is correct.
% The frame-linked reference further exposes incorrect return
% grounding behind correct revisit counts, which count-only
% evaluation cannot detect.

% \ours achieves 54.0\% micro accuracy, exceeding Spatial-MLLM~\cite{wu2025spatialmllm}, the strongest evaluated fine-tuned 3D-aware baseline, by 5.3 percentage points.
% Ablations show that removing geometric pretraining lowers micro accuracy by 4.0 percentage points, while removing trace supervision reduces performance to near chance under the tested training recipe.
% Analysis of generated traces reveals incorrect place matches and turn predictions even when final answers are correct, demonstrating the value of evaluating intermediate predictions alongside answer accuracy.
In summary, our contributions are:
\begin{itemize}[leftmargin=*,itemsep=1pt,topsep=2pt]
    \item We introduce \textbf{\bench}, a benchmark for trajectory-level spatial reasoning in egocentric walkthroughs. Its shared, frame-linked reference of places, distinct visits, and transitions supports seven question types and evaluates intermediate spatial predictions alongside final answers.
    \item We introduce \textbf{\ours}, a trace-generating multimodal reference baseline for \bench that combines appearance, geometry, and camera poses with geometric pretraining and reference-derived trace supervision.
    With a 1.7B language model, it achieves 56.6\% answer accuracy, compared with
    49.9\% for the strongest zero-shot MLLM.
    \item We evaluate general-purpose and 3D aware MLLMs and analyze intermediate reasoning traces from \ours and other baselines, revealing incorrect place or turn sequences and return-grounding errors behind correct answers.
    % \item Our evaluation quantifies the benefits of geometric and trace supervision and reveals errors in generated place matches and turn predictions even when final answers are correct.
\end{itemize}

%% file: sec/2_related.tex
\section{Related Work}
\label{sec:related}
\paragraph{Spatial reasoning benchmarks.}
Spatial question answering has been studied in reconstructed scenes~\cite{azuma2022scanqa,ma2023sqa3d}, across images~\cite{yang2026mmsibench,zhang2025spar}, and in egocentric videos~\cite{grauman2022ego4d,majumdar2024openeqa,chandrasegaran2024hourvideo}.
Video benchmarks cover tasks such as spatial measurement, motion, and route planning~\cite{yang2025thinking,li2025sti,lin2025mmsi}.
To diagnose errors beyond final-answer accuracy, MMSI-Bench~\cite{yang2026mmsibench} provides human-annotated reasoning processes for multi-image questions.
OST-Bench~\cite{lin2025ostbench} and OVO-S-Bench~\cite{li2026ovo} study streaming spatial understanding, with OVO-S-Bench additionally providing temporal evidence for revisit questions.
% Even with these evidence intervals, a correct answer alone cannot establish which visits the model identified.
\bench links reference traces of visits, paths, and turns to a shared frame--place--visit reference, enabling evaluation of intermediate spatial predictions alongside final answers (Tab.~\ref{tab:bench_compare}).
% \bench links each question's intermediate targets to the relevant video frames.
% These targets let us check which visits, paths, or turns a model recovered alongside its final answer (Tab.~\ref{tab:bench_compare}).

\input{sec/table_dataset_compare.tex}
\vspace{-4mm}
\paragraph{Spatial representations and supervision.}
Place recognition and loop closure link observations of the same location~\cite{arandjelovic2016netvlad,keetha2023anyloc,murartal2015orbslam}, while topological maps and 3D scene graphs encode spatial relationships~\cite{savinov2018sptm,chaplot2020nts,armeni20193dsg,hughes2022hydra}.
Ego-Topo~\cite{nagarajan2020egotopo} distinguishes separate visits to the same spatial zone and records transitions between zones for affordance learning and action anticipation.
\bench uses a frame-linked representation of places, distinct visits, and observed
transitions, together with metric geometry, to derive answers and reference traces for supervising and evaluating seven spatial QA tasks.
% In \bench, we link places, distinct visits, and observed transitions to video frames as a shared reference for multi-task spatial QA.
% Together with metric geometry, this reference provides answers and traces of relevant visits, paths, turns, or spatial relations across seven question types.
% These traces support both training supervision and evaluation of explicit spatial predictions, with frame links where applicable.
\vspace{-4mm}
\paragraph{3D-aware multimodal language models.}
Prior work incorporates geometry through point-cloud features~\cite{hong20233dllm,huang2024leo,chen2024ll3da}, object-centric tokens~\cite{huang2024chatscene}, 3D positional embeddings~\cite{zhu2024llava,zheng2025video3dllm}, or bird's-eye views~\cite{qi2026gpt4scene}.
Spatial-MLLM~\cite{wu2025spatialmllm}, VG-LLM~\cite{zheng2026learning}, and VLM-3R~\cite{fan2026vlm} recover geometry from RGB through feed-forward reconstruction.
Beyond answer-level training on spatial QA~\cite{chen2024spatialvlm,cheng2024spatialrgpt}, Ross3D~\cite{wang2025ross3d} uses cross-view and global-view reconstruction, while 3DRS~\cite{huang20263drs} studies correspondence supervision and
distillation from 3D foundation models.
Going a step further, \ours combines geometric pretraining with task-specific trace supervision based on \bench, which illustrates how the shared reference can train a model to output visits, paths, or turns that can subsequently be checked against the reference.
% It takes geometry features and camera poses as inputs and uses geometric prediction objectives on the language model's hidden states to learn cross-view spatial representations. 
% Fine-tuning on \bench then trains the model to describe task-relevant visits,
% paths, or turns using input frame indices before answering.

%% file: sec/table_dataset_compare.tex
\begin{table}[t]
\centering
\small
\setlength{\tabcolsep}{1.0pt}
\caption{\textbf{Comparison of spatial reasoning benchmarks.}
Revisit: repeated visits to a place; Order: temporal order of visits; Route: turn sequences along a walk;
Shared. ref.: a shared reference of place, distinct visits, and observed transitions;
Interm.\ annotations: targets for checking intermediate spatial predictions. $\checkmark$: explicit coverage; $\circ$: partial or related coverage; $\--$: not covered.}
\label{tab:bench_compare}
\begin{tabular}{lcccccc}
\toprule
Benchmark & Ego & Revisit & Order & Route & \begin{tabular}[c]{@{}c@{}}Shared\\ref.\end{tabular} & \begin{tabular}[c]{@{}c@{}}Interm.\\annot.\end{tabular} \\
\midrule
ScanQA~\cite{azuma2022scanqa}                & --         & --         & --         & --         & --         & -- \\
SQA3D~\cite{ma2023sqa3d}                     & --         & --         & --         & --         & --         & -- \\
OpenEQA~\cite{majumdar2024openeqa}           & \checkmark & --         & --         & --         & --         & -- \\
HourVideo~\cite{chandrasegaran2024hourvideo} & \checkmark & --         & --         & $\circ$    & --         & -- \\
VSI-Bench~\cite{yang2025thinking}            & \checkmark & --         & $\circ$    & $\circ$    & --         & -- \\
STI-Bench~\cite{li2025sti}              &  $\circ$   & --         & --         & $\circ$    & --         & -- \\
OST-Bench~\cite{lin2025ostbench}             & \checkmark & --         & $\circ$    & --    & --         & -- \\
MMSI-Bench~\cite{yang2026mmsibench}          & --         & --         & --         & --    & --         & $\circ$ \\
MMSI-Video-Bench~\cite{lin2025mmsi}          & $\circ$    & --   & --         & $\circ$    & --         & $\circ$ \\
SPAR-Bench~\cite{zhang2025spar}              & --         & --         & --         & --  & --         & -- \\
OVO-S-Bench~\cite{li2026ovo}                 & \checkmark & \checkmark & $\circ$    & $\circ$    & --         & $\circ$ \\
\midrule
\textbf{\bench (ours)}                       & \checkmark & \checkmark & \checkmark & \checkmark & \checkmark & \checkmark \\
\bottomrule
\end{tabular}
\end{table}

% Related benchmarks cover navigation, route planning, or trajectory descriptions, whereas our comparison emphasizes turn sequences linked to distinct visits.
% Several benchmarks derive answers from shared scene metadata or geometric annotations.
% Our reference additionally distinguishes repeated visits to the same place and records their observed transitions.

%% file: sec/3_benchmark.tex
\section{The \bench Benchmark}
\label{sec:benchmark}

\bench contains 8{,}458 multiple-choice questions from 153 SCANNet++~\cite{yeshwanth2023scannetpp} egocentric walkthroughs.
Its core is a shared, frame-linked reference of places, distinct visits, and observed transitions.
Together with metric geometry, this reference provides both answers and task-specific reference traces for training and evaluation.
For example, a \emph{revisit} trace identifies the initial visit and subsequent returns, enabling precise checks of their grounding in video frames alongside count accuracy.
% These reference information supports training supervision and allow generated spatial predictions to be checked alongside answer
% accuracy.

% 统一题型名称，明确 routing 的序列与几何依赖，区分 visit IDs 和 frame indices，并弱化未评估的潜在下游用途。否则图与精简后的正文仍会存在信息不一致。

\begin{figure*}[t]
\centering
\includegraphics[width=1.0\textwidth]{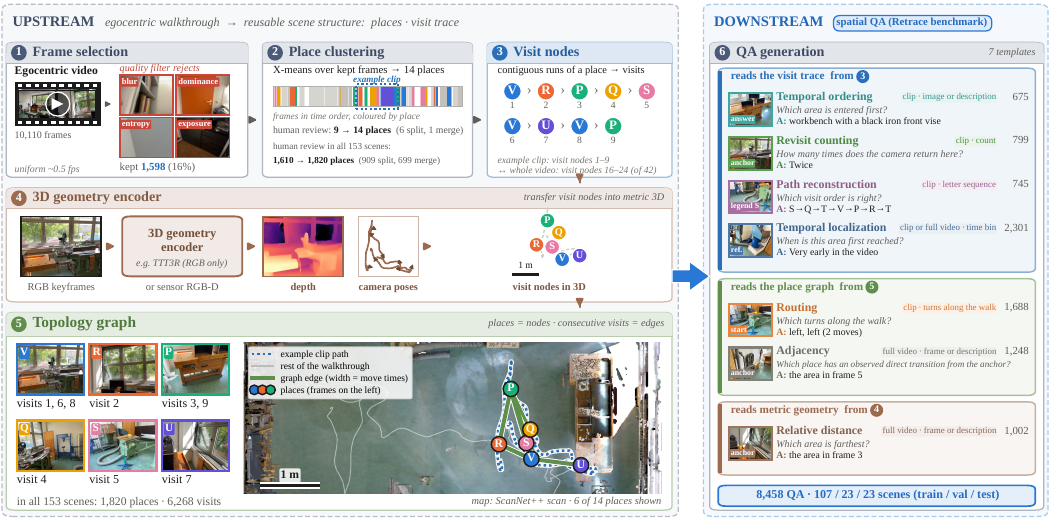}
\caption{\textbf{Constructing \bench from egocentric walkthroughs.}
We filter video frames (1), cluster retained frames into places and manually review the assignments (2), and group consecutive observations for each place into distinct visits (3).
Depth and camera poses provide metric geometry (4), while transitions between consecutive visits define a topology graph of observed connections between places (5).
These components provide answers and reference traces for seven question types (6).
In this example, place V occurs at visits 1, 6, and 8, corresponding to two returns (visit numbers here indicate the positions in the visit sequence, not frame indices).}
\label{fig:overview}
\end{figure*}
\vspace{-2mm}

\begin{figure*}[t]
\centering
\includegraphics[width=0.95\textwidth]{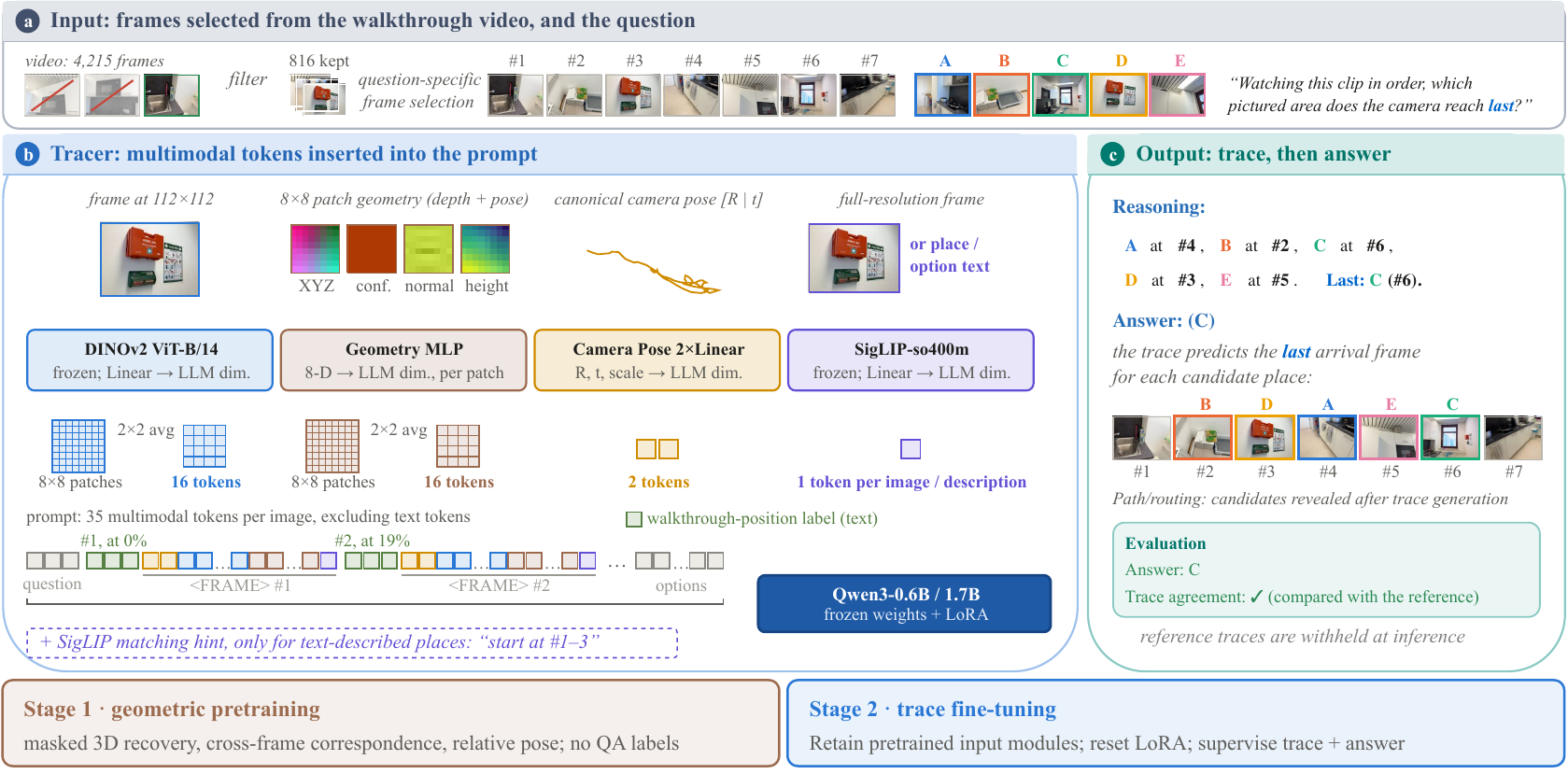}
\caption{\textbf{\ours, overview}.
(a)~Filtered video frames are selected per question. Each clip frame and image option is represented by a \texttt{<FRAME>} marker.
(b)~Each image contributes 35 multimodal tokens:
16 appearance tokens from DINOv2, 16 geometry tokens,
two camera-pose tokens, and one SigLIP token.
Walkthrough-position labels and, for text-described places, SigLIP matching hints enter as text.
(c)~\ours generates a task-specific trace followed by
an answer. The ordering example shown predicts each candidate place's last arrival frame.
For path reconstruction and routing, candidate answer sequences are provided only after trace generation.
Reference traces supervise training and support evaluation but are withheld at inference.
Bottom: Stage~1 uses geometric objectives supervision without QA labels; Stage~2 retains the pretrained input modules, initializes a fresh LoRA adapter, and further supervises traces and answers.}
\label{fig:model}
% \vspace{-4mm}
\end{figure*}

\subsection{A frame-linked reference}
\label{sec:bench_graph}

In an egocentric walkthrough, the same place may appear from different viewpoints and at different times.
To check whether a model connects these observations correctly, we record both which observations depict the same place and how they form separate visits.

% Annotators distinguish places using their spatial locations and surrounding scene layouts, rather than viewing direction alone.
We therefore distinguish a \emph{place}, a local area observed across different viewpoints, from a \emph{visit}, a maximal consecutive run of observations assigned to that place in the filtered, place-assigned frame sequence (Fig.~\ref{fig:overview}, steps 2--3).
The unassigned frames are omitted before visits are formed, and the visits therefore reflect the retained observations.
The departures and returns without sufficient retained visual evidence may not be represented.
We also provide the filtered frame sequences as video clips for inspecting the reference and model predictions.
% After unassigned frames are omitted, each change in place label begins a new visit.
For example, as shown in Fig.~\ref{fig:overview}, place $V$ appears at visits 1, 6, and 8, corresponding to two returns after the initial visit.
Place identities connect observations across views, while the visit sequence preserves their temporal order and repetition.

The visit sequence also captures transitions between places.
We construct a topological graph whose nodes represent places and whose edges record transitions between consecutive visits
(Fig.~\ref{fig:overview}, step 5).
% We connect consecutively visited places in a graph, where each edge records a direct transition observed in the walkthrough.
Each place is linked to its visits and supporting frames, while depth and camera poses provide metric positions and headings.
% Depth and camera poses provide metric positions and camera headings, while frame indices link place assignments and visits to their supporting observations.
All of these components support a shared reference for deriving answers and task-specific intermediate reference traces.
  
To construct this reference, we use ScanNet++~\cite{yeshwanth2023scannetpp} walkthroughs, following the pipeline in Fig.~\ref{fig:overview}.
After frame filtering and clustering, two annotators independently review all 153 scenes, splitting clusters that contain distinct places and merging clusters that depict the same place.
The annotators agree on all merge decisions in 82.7\% of scenes, with disagreements resolved through discussion.
Overall, 909 clusters are split and 699 clusters are merged, yielding 1{,}820 places and 6{,}268 visits.
Construction procedures and annotation details are provided in Sec.~\ref{sec:supp_bench}.

\subsection{Question types and reference traces}
\label{sec:bench_types}
\input{sec/table_question_type}
Using the shared reference, we generate questions across seven types, each with an answer and a trace of the relevant intermediate facts.
These questions and reference answers are defined over the supplied observations rather than unobserved portions of the original video.
Specifically, five trajectory-level question types concern the sequence of visits:
\textit{\textbf{temporal ordering}} compares the order of first or last visits to specified places;
\textit{\textbf{temporal localization}} identifies when a specified place is first reached;
\textit{\textbf{revisit counting}} counts returns to the starting place;
\textit{\textbf{path reconstruction}} recovers the ordered sequence of visited places, including repeated visits; and \textit{\textbf{routing}} recovers the turns taken along the recorded walk.
Two pairwise types assess relations between places: \textit{\textbf{adjacency}} asks which places have an observed direct transition from an anchor, and \textbf{\textit{relative distance}} asks which candidate is nearest to or farthest from an anchor, or whether candidates are equidistant, using Euclidean distances between their 3D reference positions.
Table~\ref{tab:qtypes} summarizes the question types and counts.
In 47\% of questions, places are specified through text descriptions, requiring models to ground them in the video.

Reference traces specify task-relevant intermediate targets: visit order, arrival frames,
return visits, place sequences, turn sequences, or pairwise spatial relations, which serve as fine-tuning targets and evaluation references.
Scoring rules and task-specific formats are provided in Sec.~\ref{sec:exp} and Sec.~\ref{sec:supp_traces}, respectively.
% Trace formats and evaluation details appear in~\ref{sec:exp} and Secs.~\ref{sec:supp_traces}.
% reference traces record the intermediate facts that support each answer: frame-to-place matches, distinct visits, paths, or turns.
% Relative-distance traces instead provide coarse geometric judgments without frame references.
% These traces supervise fine-tuning (Sec.~\ref{sec:method_finetune}) and support checking intermediate predictions alongside final answers, revealing incorrect place sequences or turns even when the selected answer is correct.
% Evaluation details and trace formats are provided in Secs.~\ref{sec:exp} and~\ref{sec:supp_traces}, respectively.
% For example, a revisit-counting trace identifies the input frames corresponding to the initial visit and subsequent returns.
% Relative-distance traces instead provide coarse near/mid/far judgments without frame references.

\subsection{Evaluation protocol and dataset statistics}
\label{sec:bench_design}

\paragraph{Question inputs.}
Each question provides indexed video frames, a prompt, and five answer options (four for
revisit counting), with places specified through images or text.
Questions and reference answers are defined over the supplied, filtered observations.
Reference traces, place assignments, visit labels, and the transition graph are not provided to models at test time.
\vspace{-4mm}
\paragraph{Shortcut checks.}
We balance answer letters, randomize place-legend letters, and select reference and option images from timestamps different from those of the input frames.
Majority-answer and option-text baselines achieve 21.6\% and 21.9\% accuracy, respectively, compared with 20.5\% chance (Tab.~\ref{tab:main}), suggesting limited predictive power from answer priors and option text alone.
\vspace{-4mm}
\paragraph{Quality checks.}
In an author-led, reference-hidden audit of 140 stratified questions from ScanNet++ and
Ego4D, answers agreed with the references on 95.7\% of questions.
Affected test questions were corrected and re-evaluated, and training checkpoints were retained (details in Sec.~\ref{sec:supp_v2}).
Details and limitations are provided in Sec.~\ref{sec:supp_audit}.
\vspace{-4mm}
\paragraph{Statistics and splits.}
\bench contains 8{,}458 question instances, with the five trajectory-level question types accounting for 73\% of the total.
We use a scene-disjoint split of 107 training, 23 validation, and 23 test scenes, containing
5{,}573, 1{,}480, and 1{,}405 questions, respectively.
Four test questions with ambiguous references are excluded from evaluation, leaving 1{,}401 test questions.
For primary results, geometric pretraining excludes all 46 validation and test scenes of \bench.
% Geometric pretraining uses either all scenes or excludes validation and test scenes, as specified in Sec.~\ref{sec:exp_ablation}.

%% file: sec/table_question_type.tex
% \begin{table}[t]
% \centering
% \small
% \setlength{\tabcolsep}{2.4pt}
% \caption{\textbf{Question types.} Input: Seg, a clip of the walkthrough; Full, frames over the whole video; Cur, a clip plus one image per involved place; ref., a reference image. Text: questions referring to places by description.}
% \label{tab:qtypes}
% \begin{tabular}{llrrr}
% \toprule
% Type & Input & All & Text & Test \\
% \midrule
% Temporal ordering     & Seg + options   & 675     & 291     & 115 \\
% Temporal localization & Seg/Full + ref. & 2{,}301 & 1{,}111 & 362 \\
% Revisit counting      & Seg             & 799     & 391     & 137 \\
% Routing               & Seg             & 1{,}688 & 824     & 250 \\
% Path reconstruction   & Seg + legend    & 745     & 348     & 124 \\
% \midrule
% Adjacency             & Cur             & 1{,}248 & 587     & 254 \\
% Relative distance     & Cur             & 1{,}002 & 458     & 163 \\
% \midrule
% Total                 &                 & 8{,}458 & 4{,}010 & 1{,}405 \\
% \bottomrule
% \end{tabular}
% \end{table}

\begin{table}[t]
\centering
\small
\setlength{\tabcolsep}{2.4pt}
\caption{\textbf{Question types and counts.}
Visual input: Seg, frames from a video segment; Full, frames over the whole video;
% Cur, a clip plus one image per involved place; 
ref., reference image (replaced by descriptions in text-phrased questions where applicable).}
\label{tab:qtypes}
\begin{tabular}{llrrr}
\toprule
Type & Visual Input & All & Text & Test \\
\midrule
Temporal Ordering
  & Seg   & 675     & 291     & 115 \\
Temporal Localization
  & Seg/Full + ref. & 2{,}301 & 1{,}111 & 362 \\
Revisit Counting
  & Seg             & 799     & 391     & 137 \\
Routing
  & Seg             & 1{,}688 & 824     & 250 \\
Path Reconstruction
  & Seg + legend    & 745     & 348     & 124 \\
\midrule
Adjacency
  & Seg + refs.             & 1{,}248 & 587     & 254 \\
Relative Distance
  & Seg + refs.             & 1{,}002 & 458     & 163 \\
\midrule
Total
  &                 & 8{,}458 & 4{,}010 & 1{,}405 \\
\bottomrule
\end{tabular}
\vspace{-4mm}
\end{table}

%% file: sec/4_method.tex
\section{\ours}
\label{sec:method}

Trajectory-level spatial reasoning requires associating observations across viewpoints while distinguishing separate visits to the same
place.
We introduce \ours, a trace-generating multimodal baseline that combines appearance, geometry, and camera poses (Fig.~\ref{fig:model}).
Geometric pretraining (Stage~1) learns input representations for spatial prediction, followed by fine-tuning (Stage~2) with \bench's reference traces to generate task-relevant spatial predictions before answering.
% Based on the thought within \bench, we introduce \ours, a lightweight multimodal model that combines geometric pretraining with reference-derived trace supervision (Fig.~\ref{fig:model}).
% Geometric pretraining trains the input modules to encode spatial information that the language model can use for 3D prediction.
% After that, trace fine-tuning supervises task-relevant place matches, visit histories, and routes using the \bench's shared reference, which also supports checking generated predictions.
% Reference trace annotations are withheld as evaluation inputs.

\subsection{Geometry-grounded inputs}
\label{sec:method_tokens}

\textbf{Cross-view representations.}
% Each frame is paired with a depth map and a scan-aligned camera pose, with geometry expressed in a shared, normalized scene coordinate system that can provide a consistent geometric reference across viewpoints.
Given video frames with depth maps and camera poses, we express depth-derived 3D positions and camera poses in a shared, normalized scene coordinate system.
Within this system, we construct patch-level geometry descriptors containing mean 3D position, depth confidence, surface normal, and height above the floor, paired with appearance features extracted by a frozen DINOv2 ViT-
B/14~\cite{oquab2023dinov2} from $112\times112$ frames.
% Learned projections encode appearance, geometry, and camera pose as separate token streams, with appearance and geometry aligned at corresponding image regions.
We project these patch-aligned features and camera poses into separate token streams and insert them at each frame's \texttt{<FRAME>} marker in the prompt.
Each frame contributes 35 tokens: 16 appearance tokens, 16 geometry tokens, two camera-pose tokens, and one SigLIP token.
% Each frame also a projected SigLIP image embedding computed at full resolution.
In addition, each input frame is preceded by a walkthrough-position label, such as ``at 64\% of the video'', indicating its relative temporal position.
The resulting sequence is processed by lightweight Qwen3 models~\cite{qwen3} (0.6B and 1.7B), with frozen
base weights and trainable LoRA adapters~\cite{hu2022lora}.
% Tokenization details are in Sec.~\ref{sec:supp_input}.
\vspace{-4mm}
\paragraph{Text-described places.}
For places specified through text, a frozen SigLIP-so400m~\cite{zhai2023siglip} matcher adds hints identifying likely matching frames, such as ``start at \#1 and \#7; destination at \#10 and \#14.''
Hints are computed from question descriptions and full-resolution input images, without reference place assignments or visit labels.
% Alongside this hint, 
% projected SigLIP image and text embeddings provide additional input tokens.
% Both image embeddings and matching hints are computed using full-resolution input images.
% For places specified through text, a frozen SigLIP-so400m~\cite{zhai2023siglip} matcher supplies a hint listing predicted frame matches, such as ``start at \#1--3; destination at \#20.''
% The matcher uses only question descriptions and input images; its matches are predictions, not reference place or visit annotations.
% Projected SigLIP image and text embeddings also enter the model. SigLIP image embeddings and matching hints use full-resolution inputs.

\subsection{Geometric pretraining}
\label{sec:method_pretrain}

% The Stage~1, pretraining stage, trains the input modules using geometric targets derived from depth and camera poses, without questions, answers, place assignments, or visit labels.
Stage~1 jointly trains the input modules using geometric targets
derived from depth and camera poses, without questions, answers, place assignments, or visit labels.
Prediction heads attached to the language model's final-layer hidden states provide supervision for three tasks under causal attention:
\textbf{\emph{masked geometry recovery}} predicts the shared-coordinate 3D positions of masked patches, weighted by depth confidence;
\textbf{\emph{cross-frame correspondence}} uses a contrastive objective to associate patches observing the same surface in adjacent frames; and
\textbf{\emph{relative-pose prediction}} estimates rotation and translation between the first and last frames with their pose tokens hidden. 
% Shared-frame geometry remains visible and can carry camera-motion information.
% The three losses are summed, with full definitions and training settings in Sec.~\ref{sec:supp_pretrain}.
Shared-coordinate geometry remains visible in the pose task, allowing the model to infer camera motion from the geometric inputs.
We jointly optimize the input modules, prediction heads, and a LoRA adapter for pretraining using the sum of the losses for three pretraining tasks (details in Sec~\ref{sec:supp_training}).

Stage~2 retains the pretrained appearance projection, geometry and pose tokenizers, and associated input embeddings, discards the prediction heads, and initializes a fresh LoRA adapter.
Thus, Stage~1 transfers the input representations learned
through geometric supervision.
% Objective definitions and training details are in Sec.~\ref{sec:supp_pretrain}.
% We additionally report a full-scene variant that includes these scenes to examine the effect of geometric exposure to evaluation environments
% (Sec.~\ref{sec:exp_ablation}).
% Stage~2 inherits the pretrained appearance projection, geometry and pose tokenizers, and associated input embeddings.
% We discard the prediction heads and replace the Stage-1 LoRA adapter with a fresh adapter.
% Thus, Stage~1 transfers input representations learned through geometric supervision within the language model.
% Default pretraining includes unlabeled frames, depth, and frame-level camera poses.
% We also report a variant excluding all 46 validation and test scenes to distinguish unseen-scene transfer from the benefit of geometric exposure (Sec.~\ref{sec:exp_ablation}).

\subsection{Frame-referenced trace supervision}
\label{sec:method_finetune}

\paragraph{Supervising places and visits.}
Stage~2 jointly fine-tunes the input modules and a fresh LoRA adapter using \bench's reference traces.
Targets take the form \texttt{Reasoning: <trace> Answer: (X)}, with next-token cross-entropy applied only to trace and answer tokens.
The trace records task-relevant place matches, visits, paths, or turns, using frame indices within the supplied clip where applicable.
Separate visits to the same place are represented by separate frame intervals.
For example, an initial visit at \texttt{\#1--3} and a later return at \texttt{\#20} are listed separately, supervising not only how many times the place is visited but also which observations support each visit.
Relative-distance traces instead provide coarse distance judgments without frame references.
Formats for all seven question types appear in Sec.~\ref{sec:supp_traces}.
% Targets are derived from the shared reference in
% Sec.~\ref{sec:benchmark} and describe frame-to-place matches, visits, paths, or turns.
% Frame indices identify slots within the supplied clip.
% Reference visit boundaries are preserved through frame-indexed intervals: an initial visit at \texttt{\#1--3} and a return at \texttt{\#20} after visiting another place remain separate
% entries for the same place.
% Relative-distance traces instead provide coarse geometric judgments without frame references.
% Formats for all seven question types appear in
% Sec.~\ref{sec:supp_traces}.

At evaluation, reference traces, place assignments, visit annotations, and the transition graph are withheld.
The model generates its own trace and answer, which are scored separately for spatial agreement with the reference and
final-answer correctness.
% This can reveal incorrect trace grounding even when the final answer is correct.
\vspace{-4mm}
\paragraph{Predicting paths before seeing answer options.}
For routing and path reconstruction, we withhold answer options until the trace is complete, enabling sequence recovery to be evaluated without access to those candidates.
The initial prompt contains the question, clip frames, and any place-reference or legend images.
During training, options are inserted between the reference trace and the target answer.
At inference, the model first generates its trace, then receives the options and selects an answer with the generated trace retained in context.
This prevents the trace from directly copying an option, while allowing the final answer to differ from the predicted sequence.
For other question types, answer options are visible from the start, but the model still generates a trace before answering.
% Other question types include candidates in the initial prompt.

% \paragraph{Optimization.}
% We jointly fine-tune the transferred input modules and a fresh LoRA adapter on the \bench training split.
% Next-token cross-entropy supervises trace and answer tokens, excluding supplied candidate text.
% Answer options are randomly permuted during training.
% Training and decoding details appear in
% Sec.~\ref{sec:supp_finetune}.

%% file: sec/5_experiment_v2.tex
\section{Experiments}
\label{sec:exp}

% We first evaluate current MLLMs and \ours on \bench (Sec.~\ref{sec:exp_main}), examine sequence and visit-grounding errors hidden by answer accuracy, and analyze \ours's components and transfer to Ego4D (Sec.~\ref{sec:exp_analysis}).

\subsection{Setup}
\label{sec:exp_setup}

\paragraph{Dataset and metrics.}
We use the scene-disjoint split described in Sec.~\ref{sec:bench_design}: 107 training, 23 validation, and 23 test scenes.
% In the primary setting, Stage~1 excludes all validation and test scenes; the \emph{full-scene} setting also uses their unlabeled RGB-D frames and poses, and serves as additional analysis.
We report answer accuracy over questions (Micro) and
averaged across the seven task types (Macro).
% Reported differences use 95\% paired bootstrap intervals over test scenes; these intervals do not capture training-seed variation.\\
% Micro is accuracy over questions; Macro averages the seven task accuracies, and Traj.\ and Pair.\ average their task groups.
% Differences between models come with 95\% intervals (in brackets) from a paired bootstrap over test scenes; an interval that includes zero does not establish equal performance, and scene resampling does not capture training-run variation.
\vspace{-4mm}
\paragraph{Baselines.}
We evaluate nine zero-shot general-purpose MLLMs, three 3D-aware model families, and option-only controls (Tab.~\ref{tab:main}).
Each 3D-aware baseline receives its required input modalities.
The fine-tuned baselines use answer-letter supervision and the same number of training-question presentations as \ours.
Model-specific inputs are summarized in Tab.~\ref{tab:inputs}.
% \ours additionally uses reference trace supervision, sensor geometry, and a SigLIP matching hint.
% Like the 3D-aware baselines but unlike the zero-shot API models, it receives walkthrough-position labels, which give each frame's relative time in the video (\eg, ``at 64\% of the video''); they bear directly on temporal localization.
For path reconstruction and routing, \ours generates traces before seeing answer options, whereas the baselines and our letter-only controls see them from the start.
Comparisons therefore reflect complete systems that differ in input information, supervision, and answer-option timing.
% Answer accuracy therefore compares complete systems, including their inputs and inference protocols, and does not isolate any single component (inputs per model in Tab.~\ref{tab:inputs}).
% Frames are $112\times112$ unless noted in Tab.~\ref{tab:main}; \ours's SigLIP inputs use full-resolution frames.
% \paragraph{Configuration selection.}
% The configuration of \ours and all checkpoints were selected on validation before the first test evaluation.
% Afterwards, four alternatives were evaluated on the test scenes (SigLIP from 112-px frames, an image--image matching hint, a heading feature, and scene memory), and the decision to keep the original configuration rather than the heading variant was made after that variant's second test seed (supplement).
% The test set is therefore not untouched for the choice of reported configuration, which affects every reported \ours row, and the component comparisons of Sec.~\ref{sec:exp_ablation} are exploratory; the Ego4D evaluation (Sec.~\ref{sec:exp_ego4d}) is the only one made after the configuration was frozen, and it is a partial check (different task mix, unaudited references).
\vspace{-4mm}
\paragraph{Implementation details.} 
Stage~1 runs for 5k steps with 128 clips per step and a peak learning rate of $2.8{\times}10^{-4}$.
Stage~2 and fine-tuned baselines use 6k steps with an effective batch size of 4, at learning rates of $2{\times}10^{-4}$ and $2{\times}10^{-5}$, respectively.
Checkpoints are selected on validation data, and all models are trained on a single NVIDIA RTX A6000 (details in Sec.~\ref{sec:supp_impl}).
% In the first pretraining stage, we train for 5k steps with 128 clips per step with a peak learning rate of $2.8{\times}10^{-4}$.
% For the fine-tuning stage, we trains for 6k steps with a batch of 4 questions, and a constant learning rate of $2{\times}10^{-4}$, the checkpoint is chosen on the validation sets. 
% Fine-tuned base use the same number of steps and question presentations at learning rate $2{\times}10^{-4}$, and all models are trained on a single NVIDIA RTA A6000, full settings are in Sec. ~\ref{sec:supp_impl}.

\subsection{Main results}
\label{sec:exp_main}

\input{sec/table_main_v2}

\paragraph{Benchmark difficulty.}
As shown in Tab~\ref{tab:main}, the option-text-only classifier and training-prior majority baseline achieve 21.6\% and 21.3\% micro accuracy, respectively, both close to chance (20.5\%). These results suggest that the tested
shortcuts provide little advantage over chance.
The best zero-shot model, Claude Opus 5.5, answers 49.9\% Micro accuracy, while revisit counting, routing, and adjacency remain challenging for it.
\ours achieves the highest observed overall answer accuracy
under its richer input and supervision setting, while
performance varies substantially across tasks.
% With a 1.7B language model, \ours achieves a competitive
% 55.8\% micro accuracy, supporting the practical value of
% training a compact model with reference-derived intermediate
% traces.
% We next examine these intermediate predictions to determine whether correct answers are supported by correct spatial facts.
\vspace{-4mm}
\paragraph{In-domain fine-tuning.}
The 3D-aware models Spatial-MLLM and Video-3D LLM achieve
only 22.3\% and 17.1\% micro accuracy zero-shot, respectively.
Fine-tuning on the \bench training split raises Spatial-MLLM to 49.1\%, with the largest gains on temporal localization and routing.
Using VLM-3R's tower and fusion block, the fine-tuned LLaVA-OneVision-0.5B and Aquila-VL-2B backbones reach 32.2\% and 49.1\%, respectively.
However, fine-tuned Video-3D LLM remains near chance at 21.6\%, indicating that in-domain training alone does not ensure strong performance across architectures.
\vspace{-4mm}
\paragraph{\ours performance.}
With a 1.7B language model, \ours achieves the highest macro and micro accuracy among evaluated models (56.3\% and 56.6\%), exceeding all evaluated fine-tuned and zero-shot baselines.
Meanwhile, the smaller 0.6B variant remains competitive at 51.6\% micro accuracy and achieves the highest revisit-counting accuracy (48.9\%).
Separately, allowing Stage~1 to use unlabeled frames, depth, and poses from evaluation scenes raises the 0.6B model's micro accuracy to 54.2\%.
However, the performance of \ours remains uneven across tasks: \ours achieves the highest accuracy on temporal ordering
(81.7\%) and localization (79.8\%), but trails the
strongest baselines on routing (34.4\%) and path
reconstruction (55.6\%).
These task-level differences motivate our further analysis of intermediate spatial predictions.
\vspace{-4mm}

\begin{figure*}[t]
\centering
\includegraphics[width=\textwidth]{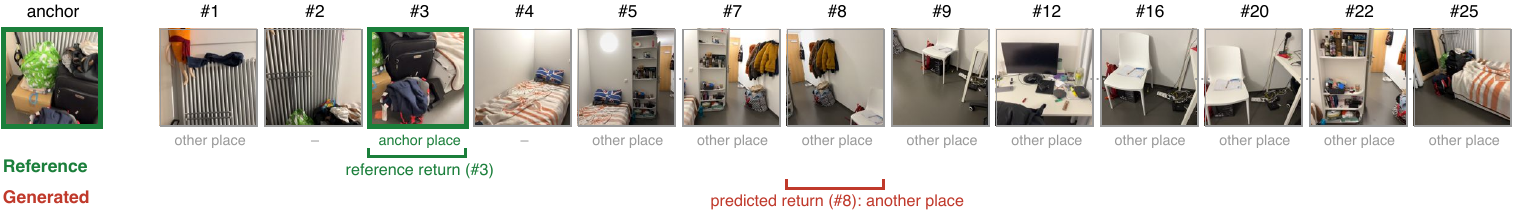}
\caption{\textbf{A correct revisit count with reference-disagreeing grounding}. Asked how many more times the camera returns to the anchor place, \ours answers ``Once'', which is correct, but its trace puts the return at \#8, another place, and misses the reference return at \#3.}
\label{fig:trace_grounding}
\end{figure*}
\paragraph{Training and component analysis.}
As shown in Table~\ref{tab:main}, geometric pretraining yields a modest overall improvement whose scene-bootstrap interval includes zero.
In addition, walkthrough-position labels substantially affect performance: removing them reduces Micro accuracy from 51.6\% to 45.6\% (supplement, Tab.~\ref{tab:abl_small}).
These results indicate a relative task-specific benefit of geometric pretraining and the model's dependence on position inputs.
Full analyses and ablation studies are provided in Sec~\ref{sec:exp_ablation} and ~\ref{sec:supp_ablations}.
% \vspace{-2mm}
\subsection{What answer accuracy misses}
\label{sec:exp_analysis}

\input{sec/table_acc_dig}
Table~\ref{tab:trace_diag} evaluates primary \ours
(0.6B; Stage~1 excludes evaluation scenes) and zero-shot
Gemini 3.1 Flash-Lite~\cite{google2026gemini31flashlite}, prompted in two turns to generate a sequence before receiving answer options.
We report reference agreement, mismatch among correct answers, and sequence mismatch after allowing one edit.

The shared reference supports two complementary diagnostics.
First, \emph{independent sequence recovery} tests reconstruction without candidate sequences. 
Among \ours's correct answers, 69\% of paths and 61\% of routes have incorrect generated sequences, remaining 50\% and 36\% after allowing one edit.
Gemini exhibits the same gap: 71\% of its correct path answers and 76\% of its correct routing answers accompany incorrect sequences.
These mismatches indicate a gap between sequence recovery and answer selection: models may revise their predictions after seeing the candidates, so an incorrect trace does not necessarily imply an incorrect final decision.

Second, \emph{visit grounding} checks whether predicted returns correspond to the correct observations. 
Seven of \ours's 65 correct, count-consistent revisit
answers disagree with the reference grounding.
A reference-blind review identifies three clear wrong-place errors, and most others concern visit boundaries.
Figure~\ref{fig:trace_grounding} shows a correct return count
grounded at another place, missing the actual return.
Unlike sequence comparison against the correct option,
this check requires our reference's frame--place--visit links.

\subsection{Evaluation on Ego4D Walkthroughs}
\label{sec:exp_ego4d}

\input{sec/table_ego4d}

We apply the same place--visit representation, question generator, and reference-trace format to 40 Ego4D walkthroughs~\cite{grauman2022ego4d}, producing 955 questions across all seven task types.
Unlike \bench's sensor geometry, depth and poses are estimated using TTT3R~\cite{chen2026ttt3r}.
Annotators review places and visits using the same procedure as for \bench.
This evaluation examines both the applicability of our dataset construction pipeline and the cross-dataset transfer of \ours.
As shown in Tab~\ref{tab:ego4d}, \ours achieves the highest micro accuracy on this set: 56.0\% with the 1.7B language model and 54.7\% with the 0.6B model,
without additional training or configuration selection
on Ego4D.
\ours achieves the highest Micro accuracy (56.0\% at 1.7B; 54.7\% at 0.6B), while on the other six task types, both variants trail Claude Opus 5.5 by about 10 percentage points
(Sec.~\ref{sec:supp_exp}).
These results support the applicability of the construction pipeline and cross-dataset transfer abilities for \ours despite changes in video source and geometry estimation.

%% file: sec/table_main_v2.tex
% 2026-11-16: numbers from release_archive/v2_corrected/tables/results_tables_v2.tex (Retrace v2). Earlier: copied from ../results_tables.tex (generated by make_results_tables.py, 2026-11-12: primary / scaling / full-scene blocks; Aquila VLM-3R row); re-copy after regenerating
\begin{table*}[t]
\centering
\small
\setlength{\tabcolsep}{3.2pt}
\caption{\textbf{Results on the \bench\ test scenes} (1{,}401 questions' accuracy, \%).
Ord.: temporal ordering; Loc.: temporal localization; Rev.: revisit counting; Rte.: routing; Pth.: path reconstruction; Adj.: adjacency; Dst.: relative distance;
Traj.\ and Pair.\ average their types, Macro averages all seven, Micro is over all questions. FT: fine-tuned on our training split; ZS: zero-shot.
Frames are 112$\times$112 except $^\dagger$512\,px and $^\ast$native resolution (Spatial-MLLM does not learn from 112\,px).
$^a$Picks the option whose text was most often correct for the type in training; $^b$a classifier trained on the option texts only (no video).
$^c$\ours's SigLIP embeddings are otherwise computed from the full-resolution frames. \ours rows without $^d$: Stage~1 pretrained without the 46 validation and test scenes; $^d$Stage~1 also saw the unlabeled frames, depth, and poses of these scenes (no questions or labels).
$^\star$VLM-3R's CUT3R tower and fusion block on LLaVA-OneVision-0.5B~\cite{li2024llava} or Aquila-VL-2B~\cite{gu2024infinitymm}, fine-tuned with the same setting.
\textbf{Bold}: best; \underline{underlined}: second best among the baselines and \ours\; the full-scene block is not ranked.}
\label{tab:main}
\begin{tabular}{llc ccccc c cc c c c}
\toprule
& Model & LLM & Ord. & Loc. & Rev. & Rte. & Pth. & Traj. & Adj. & Dst. & Pair. & Macro & Micro \\
\midrule
& Chance & -- & 20.0 & 20.0 & 25.0 & 20.0 & 20.0 & 21.0 & 20.0 & 20.0 & 20.0 & 20.7 & 20.5 \\
& Majority (training prior)$^a$ & -- & 25.2 & 27.6 & 31.6 & 15.6 & 11.3 & 22.3 & 16.1 & 20.9 & 18.5 & 21.2 & 21.3 \\
& Text only (options)$^b$ & -- & 21.7 & 18.0 & 31.6 & 23.6 & 19.4 & 22.8 & 20.9 & 20.9 & 20.9 & 22.3 & 21.6 \\
\midrule
\multirow{10}{*}{\rotatebox{90}{\scriptsize Zero-shot}} 
% & Gemini 3.8 Flash & -- & 68.7 & 51.4 & 36.8 & 47.2 & 87.9 & 58.4 & 39.0 & 65.6 & 52.3 & 56.7 & 53.3 \\
& GPT-6 Luna~\cite{openai2026gpt6luna} & -- & 69.6 & 51.1 & 30.8 & 43.2 & 67.7 & 52.5 & 33.5 & 47.2 & 40.4 & 49.0 & 47.1 \\
& Claude Opus 5.5~\cite{anthropic2026opus55} & -- & \underline{73.9} & 52.2 & 33.1 & 33.6 & \textbf{79.8} & 54.5 & \underline{36.6} & \textbf{64.4} & \textbf{50.5} & \underline{53.4} & 49.9 \\
& Claude Sonnet 5.5~\cite{anthropic2026sonnet55} & -- & 68.7 & 45.6 & 34.6 & 33.6 & \underline{78.2} & 52.1 & 31.9 & 50.3 & 41.1 & 49.0 & 45.3 \\
& Claude Haiku 5.5~\cite{anthropic2026haiku55} & -- & 60.9 & 40.9 & 35.3 & 24.4 & 70.2 & 46.3 & \underline{36.6} & 45.4 & 41.0 & 44.8 & 41.4 \\
& Gemini 3.5 Flash-Lite~\cite{google2026gemini35flashlite} & -- & 61.7 & 49.4 & 37.6 & 31.2 & 71.8 & 50.4 & 29.5 & 44.8 & 37.2 & 46.6 & 43.9 \\
& Gemini 3.1 Flash-Lite~\cite{google2026gemini31flashlite} & -- & 60.9 & 40.9 & 30.8 & 23.6 & 47.6 & 40.8 & 27.6 & 42.9 & 35.3 & 39.2 & 36.9 \\
& GPT-4o~\cite{openai2024gpt4o} & -- & 39.1 & 29.8 & 45.1 & 26.4 & 38.7 & 35.8 & 26.0 & 33.1 & 29.6 & 34.0 & 31.9 \\
& Qwen3-VL~\cite{qwen3} & 235B & 55.7 & 37.8 & 25.6 & 25.2 & 62.9 & 41.4 & 27.6 & 42.9 & 35.3 & 39.7 & 36.8 \\
& Qwen2.5-VL$^\dagger$~\cite{bai2025qwen25vl} & 7B & 38.3 & 26.8 & 25.6 & 22.8 & 26.6 & 28.0 & 22.0 & 26.4 & 24.2 & 26.9 & 26.0 \\
\midrule
\multirow{2}{*}{\rotatebox{90}{\scriptsize \shortstack{3D\\(ZS)}}} & Spatial-MLLM~\cite{wu2025spatialmllm} & 3B & 28.7 & 19.3 & 7.5 & 27.2 & 22.6 & 21.1 & 21.7 & 29.4 & 25.6 & 22.3 & 22.3 \\
& Video-3D LLM~\cite{zheng2025video3dllm} & 7B & 19.1 & 24.9 & 8.3 & 9.2 & 17.7 & 15.8 & 15.4 & 20.2 & 17.8 & 16.4 & 17.1 \\
\midrule
\multirow{4}{*}{\rotatebox{90}{\scriptsize 3D (FT)}} & Spatial-MLLM (FT)$^\ast$~\cite{wu2025spatialmllm} & 3B & 56.5 & 60.8 & 41.4 & \textbf{55.6} & 64.5 & 55.8 & 27.6 & 36.2 & 31.9 & 48.9 & 49.1 \\
& VLM-3R$^\star$ (FT)~\cite{fan2026vlm} & 0.5B & 48.7 & 37.6 & 37.6 & 39.6 & 15.3 & 35.8 & 22.8 & 20.2 & 21.5 & 31.7 & 32.2 \\
& VLM-3R$^\star$ (FT)~\cite{fan2026vlm} & 1.5B & 63.5 & 55.2 & 46.6 & \underline{48.4} & 66.1 & \underline{56.0} & 33.1 & 40.5 & 36.8 & 50.5 & 49.1 \\
& Video-3D LLM (FT)~\cite{zheng2025video3dllm} & 7B & 16.5 & 19.9 & 33.8 & 24.0 & 17.7 & 22.4 & 21.7 & 18.4 & 20.0 & 21.7 & 21.6 \\
\midrule
& \ours (ours, Qwen3-0.6B) & 0.6B & 71.3 & \underline{74.0} & \textbf{48.9} & 35.2 & 50.0 & 55.9 & 33.9 & 44.2 & 39.0 & 51.1 & \underline{51.6} \\
% \midrule
& \ours\ (ours, Qwen3-1.7B) & 1.7B & \textbf{81.7} & \textbf{79.8} & \underline{47.4} & 34.4 & 55.6 & \textbf{59.8} & \textbf{40.9} & \underline{54.0} & \underline{47.5} & \textbf{56.3} & \textbf{56.6} \\
\midrule
\multirow{3}{*}{\rotatebox{90}{\scriptsize \shortstack{Full\\scene$^d$}}} & \ours & 0.6B & 71.3 & 80.9 & 45.9 & 34.8 & 48.4 & 56.3 & 39.0 & 47.9 & 43.4 & 52.6 & 54.2 \\
 & \ours, seed 2 & 0.6B & 79.1 & 77.1 & 46.6 & 36.0 & 41.9 & 56.2 & 40.2 & 50.9 & 45.5 & 53.1 & 54.2 \\
 & \ours, all inputs at 112\,px$^c$ & 0.6B & 73.0 & 80.4 & 47.4 & 30.0 & 50.0 & 56.2 & 35.8 & 41.1 & 38.5 & 51.1 & 52.3 \\
\bottomrule
\end{tabular}
\end{table*}

%% file: sec/table_acc_dig.tex
% \begin{table}[t]
% \centering
% \footnotesize
% \setlength{\tabcolsep}{2.2pt}
% \caption{\textbf{Trace diagnostics} of \ours (0.6B, test). Agree (\%): percentage of questions whose predicted spatial fact matches the reference, , scored separately from answer correctness.
% Mismatch among correct: reference disagreement among correctly answered questions, with counts shown in parentheses. $^\dagger$An adjacent clip frame is also accepted. Rules and exclusions are detailed in Sec~\ref{sec:supp_traces}.}
% \label{tab:trace_diag}
% \begin{tabular}{@{}lllrrr@{}}
% \toprule
% Task & Fact & Rule & $n$ & Agree(\%) & \shortstack[r]{Mismatch\\among correct} \\
% \midrule
% Path & place seq. & exact & 124 & 15.3 & 69\% (43/60) \\
% Routing & turn seq. & exact & 250 & 13.6 & 61\% (54/88) \\
% Revisit & return visits & per visit & 137 & 43.6 & 11\% (7/65) \\
% Loc., window & arrival frame & visit\,$\dagger$ & 189 & 89.9 & 0\% (0/157) \\
% Loc., video & stage frame & exact & 173 & 57.8 & 12\% (13/111) \\
% \bottomrule
% \end{tabular}
% \vspace{-2mm}
% \end{table}

\begin{table}[t]
  \centering
  \footnotesize
  \setlength{\tabcolsep}{2pt}
  \caption{\textbf{Trace diagnostics on the test set.}
  Agree: reference agreement over all questions of the task.
  Mismatch among correct: reference disagreement among
  correct answers, with counts in parentheses.
  ${>}1$ edit: percentage of correct answers whose predicted
  sequence requires more than one insertion, deletion,
  or substitution to match the reference.
  $^\dagger$Same reference visit or an adjacent clip frame.
  Rules and exclusions appear in Sec.~\ref{sec:supp_traces}.}
  \label{tab:trace_diag}
  \begin{tabular}{@{}lllrrrr@{}}
  \toprule
  Task & Fact & Rule & $n$
  & \shortstack{Agree\\(\%)}
  & \shortstack{Mismatch\\among correct}
  & \shortstack{${>}1$ edit\\(\%)} \\
  \midrule
  \multicolumn{7}{@{}l}{\emph{\ours (0.6B)}} \\
  Path & place seq. & exact
  & 124 & 15.3 & 69\% (43/62) & 50 \\
  Routing & turn seq. & exact
  & 250 & 13.6 & 61\% (54/88) & 36 \\
  Revisit & return visits & per visit
  & 133 & 43.6 & 11\% (7/65) & -- \\
  Loc., window & arrival frame & visit$^\dagger$
  & 189 & 89.9 & 0\% (0/157) & -- \\
  Loc., video & stage frame & exact
  & 173 & 57.8 & 12\% (13/111) & -- \\
  \midrule
  \multicolumn{7}{@{}l}{\emph{Gemini 3.1 Flash-Lite (trace-first)}} \\
  Path & place seq. & exact
  & 124 & 19.4 & 71\% (60/84) & -- \\
  Routing & turn seq. & exact
  & 250 & 7.6 & 76\% (59/78) & -- \\
  \bottomrule
  \vspace{-4mm}
  \end{tabular}
  \end{table}

%% file: sec/table_ego4d.tex
% % generated by figures/make_ego4d_tables.py from NaviMLLM/baseline_runs/ego4d_* -- do not edit by hand
% generated by figures/make_ego4d_tables.py from NaviMLLM/baseline_runs/ego4d_* -- do not edit by hand

\input{sec/table_ego4d_test}

%% file: sec/table_ego4d_test.tex
\begin{table}[t]
\centering
\footnotesize
\setlength{\tabcolsep}{2.0pt}
\caption{\textbf{Generalization to Ego4D walkthroughs} (Ego4D-derived evaluation set: 955 questions, 40 videos; no training on it; accuracy, \%). Models retain their Retrace settings without additional training on this set; \ours uses depth and poses estimated by TTT3R. 
$^\dagger$512\,px inputs.
Bold / underline: best / second (lower bounds not ranked).}
\label{tab:ego4d}
\begin{tabular}{@{}llcccc@{}}
\toprule
Model & LLM & Loc. & Rev. & Dst. & Micro \\
\midrule
Chance & -- & 20.0 & 25.0 & 20.0 & 20.7 \\
Majority (training prior) & -- & 35.2 & 33.8 & 22.2 & 30.6 \\
Text only (options) & -- & 19.9 & 33.8 & 22.2 & 22.6 \\
\midrule
% Gemini 3.8 Flash & -- & 49.0 & \textbf{39.8} & 37.5 & 49.3 \\
Claude Opus 5.5 & -- & 44.3 & 37.6 & 34.7 & 46.1 \\
Claude Sonnet 5.5 & -- & 38.0 & 37.6 & 22.2 & 40.6 \\
Claude Haiku 5.5 & -- & 40.2 & 35.3 & 33.3 & 41.5 \\
Gemini 3.1 Flash-Lite & -- & 43.7 & 34.6 & 26.4 & 41.0 \\
GPT-4o & -- & 34.8 & \underline{38.3} & 23.6 & 34.9 \\
Qwen2.5-VL$^\dagger$ & 7B & 25.0 & 29.3 & 30.6 & 27.9 \\
Qwen3-VL$^\dagger$ & 235B & 40.7 & 27.1 & 27.8 & 39.5 \\
\midrule
Spatial-MLLM (FT) & 3B & 61.0 & 30.8 & 23.6 & 49.3 \\
VLM-3R (FT) & 0.5B & 30.5 & \textbf{39.1} & 22.2 & 31.2 \\
VLM-3R (FT) & 1.5B & 51.6 & \underline{38.3} & 30.6 & 47.1 \\
\midrule
\ours & 0.6B & \underline{69.3} & \textbf{39.1} & \underline{38.9} & \underline{54.7} \\
\ours & 1.7B & \textbf{71.5} & 34.6 & \textbf{43.1} & \textbf{56.0} \\
\bottomrule
\end{tabular}
\vspace{-4mm}
\end{table}

%% file: sec/6_conclusion.tex
\section{Conclusion}
We introduced \bench, whose frame-linked references support answer evaluation and diagnosis of trajectory-level spatial predictions.
By linking places, distinct visits, and observed transitions,
\bench provides supervision and evaluation targets across seven spatial question types.
Our compact baseline, \ours, achieves the highest observed
micro accuracy among evaluated models at 56.6\%.
Our findings show that correct answers can accompany
incorrect spatial predictions, highlighting the value
of evaluating intermediate traces alongside final answers.
Furthermore, these diagnostics identify place grounding and visit tracking as concrete targets for improving trajectory-level spatial reasoning in future multimodal models.

\section*{Acknowledgments}
We would like to thank Yurun Shi for her valuable assistance with early-stage data annotation and verification.

%% file: sec/X_suppl.tex
% Supplementary material, compiled after the main paper (author-kit convention);
% split the PDF after the references for submission.
\clearpage
\setcounter{page}{1}
\maketitlesupplementary

\section{Benchmark construction details}
\label{sec:supp_bench}

\paragraph{Frame selection and place discovery.}
Frames are sampled at about 0.5\,fps, and blurred, poorly exposed, low-entropy, and near-duplicate frames are removed.
The kept frames are clustered into places by appearance: each frame is encoded with CLIP ViT-B/32, then these embeddings are reduced to 50 dimensions with PCA, and X-means chooses the number of clusters by the Bayesian information criterion (starting from 2, at most 12 per scene). 
Consecutive frames with the same label form a visit, and runs shorter than 10 frames are treated as transitions. 
The clustering is then held to 3--31 places per scene.
At each merge step, we select the pair of temporally adjacent clusters with the highest centroid cosine similarity and merge them if that similarity is at least 0.9. For heterogeneous clusters, we use k-means with \(k=2\) to obtain two temporally contiguous parts. We keep the split only if the similarity between the two parts is at most 0.92.
Annotators then review every scene and split places that mix distinct locations and merge places that show the same one (909 splits and 699 merges over 153 scenes, 1{,}610 $\rightarrow$ 1{,}820 places, 6{,}268 visits).

\paragraph{Geometric verification.}
For each visit, we measure the distance between its camera center and that of the place’s medoid visit, or the 3D points the camera looks at. We flag the visit if one of the distances exceed 2.5 times the place’s within-visit spread.
Places with no flagged visit form Tier~A, places flagged on position only Tier~B, and places flagged on both Tier~C; over the 153 scenes, 515 places are Tier~A (698 visits), 753 Tier~B (2{,}773 visits), and 552 Tier~C (2{,}797 visits). 
Tier~C places are never the subject of a question (an option, anchor, legend, start, or destination place).
Revisit questions additionally require geometrically consistent revisits or a manual check of each qa pairs.

\paragraph{Place descriptions.}
Each Tier~A/B place is described from its two or three representative keyframes by Claude Opus~4.1~\cite{anthropic2025claudeopus41}, run as an agent with a fixed labeling procedure: a short phrase (6--12 words requested) with at least one concrete, visually distinctive detail, such as an object, color, furniture arrangement, or architectural feature, and no numbers, letters, or identifiers (\eg, ``blue steel waste barrel on a white brick pedestal by the door'').
A word-overlap check (Jaccard similarity of stemmed words above 0.3) compares every pair of descriptions within a scene, and the flagged pairs are rewritten for up to two rounds, and descriptions that still collide are not used in questions (9 of 1{,}216 clusters used in generating QA pairs).
Questions never refer to places by opaque identifiers such as ``Room~2''.

\paragraph{Question details.}
% Relative-distance winners must be clearly separated (best-to-second distance ratio below 0.85); for near-ties (ratio $\geq0.95$) ``Equidistant'' is correct, and cases in between are discarded.
For relative-distance measurement, we compute the ratio \(r\) of the smallest distance to the second-smallest distance. If \(r < 0.85\), the closest candidate is the correct answer. If \(r \ge 0.95\), the correct answer is “Equidistant.” We discard cases with \(0.85 \le r < 0.95\).
Ordering clips are padded so that the answer is not the first or last frame, and temporal clips place the arrival in a uniformly sampled stage.
Option and reference images are at least 45 frames from any input frame.
A fact that cannot support distinct options is skipped rather than padded.
With legend letters assigned in order of first visit, the correct path always listed its letters in alphabetical order of first appearance (745 of 745 path questions): the text-only rule ``pick an option whose letters first appear in alphabetical order'' reached 61.8\% (chance 20\%), and a text-only classifier that combines it with other option-text cues solved 99.7\%. 
After randomizing the assignment, the correct path is alphabetical in 14.5\% of the questions and the rule scores 21.0\% on the final corpus.

\section{Traces for all question types}
\label{sec:supp_traces}

\begin{itemize}[leftmargin=*,itemsep=1pt]
\item \emph{Temporal localization}: the clip frame that shows the place (``Context frame \#3.'').
\item \emph{Temporal ordering}: the frame of each option's place and the earliest (``A at \#12, B at \#4, \dots{} First: D (\#1).'').
\item \emph{Revisit}: the anchor's stays and the number of returns (``Anchor room: \#1--2 (start). Returns: 0.'').
\item \emph{Path reconstruction}: the frames of each legend place and the path (``P not seen, Q at \#5, R at \#11--13, S at \#1--3. Path: S$\to$Q$\to$R.'').
\item \emph{Routing}: the start, the turn at each move, and the destination (``Start at \#1--2. \#5--12: turn left. Destination at \#13--14: go straight.'').
\item \emph{Adjacency}: the frames of the anchor and of each option's place, and which option is entered from the anchor (``Entered from anchor: A (\#8).'').
\item \emph{Relative distance}: a near, mid, or far judgment per option (no frame references) and the nearest (``A: far. C: near. \dots{} Nearest: C.'').
\end{itemize}

\paragraph{Trace evaluation (Sec.~\ref{sec:exp_analysis}).}
This section details how the generated traces analyzed in Sec.~\ref{sec:exp_analysis} (Tab.~\ref{tab:trace_diag} and Tab.~\ref{tab:trace_diag_full}) are parsed and scored against the reference traces.

The generated trace is the text between ``Reasoning:'' and ``Answer:''; the answer is the last parenthesized letter of the output, so trace correctness is judged independently of it.
Generated and reference traces are parsed into their spatial content and compared field by field: 
\begin{itemize}
    \item \textit{Temporal ordering.} The decisive option named on the \textit{First} or \textit{Last} line.
    \item \textit{Temporal localization.} The first-arrival frame.
    \item \textit{Revisit.} The anchor's stay intervals
    \item \textit{Distance.} The near, mid, or far label assigned to each option (these traces do not reference frames).
    \item \textit{Path reconstruction.} The letter sequence after ``Path:''; routing, the ordered turn words.
\end{itemize}
The parser extracts these fields from all generated and reference traces of the seven types. Adjacency is evaluated on the 156 questions that ask where the camera goes from the anchor (the entered option's letter); the 98 that ask where it never goes have reference traces that list the anchor's neighbors and name no entered place.

Parsed and whole-text comparison agree on 100\% of ordering and localization traces, 98\% of revisit, 96\% of distance, 79\% of path, and 83\% of routing traces (full-scene model); the disagreements are wording differences that do not change the parsed content (\eg, ``Start at start frame'' vs.\ ``Start at \#1--2'').

\textbf{\emph{Visit-level agreement.}} A different frame of the same visit is an equally valid grounding. We map every clip frame to its place and visit with the reference (its keyframe label, else the visit whose frame range contains it) and count a generated frame as agreeing if it lies in the same visit as the reference frame, and in windowed localization also if it is the adjacent clip frame (transition frames between labelled visits belong to no visit).
% revisit return stays agree if their number matches and each overlaps the corresponding reference stay (the anchor's starting stay, which some traces list as ``(start)'', is ignored). 

For revisit counting, predicted and reference returns are matched in temporal order.
Agreement requires the same number of returns, with each
predicted interval overlapping the corresponding reference
visit; an unlabelled frame immediately adjacent to that
visit is also accepted.
The initial stay is excluded from both lists.
As shown in Table~\ref{tab:trace_diag_full}, across the 137 revisit questions, all 170 reference stays (128 returns and 42 starting stays) lie on the anchor place, and 184 of 189 windowed localization reference frames show the target place. 

In full-video localization, the clip is a set of stage frames and the trace names the stage frame nearest the first arrival, which need not show the place, so only exact agreement applies (an adjacent frame there is a different stage). 
% For ordering we compare the decisive first/last option; the per-option frames enter only the partial score.

\textbf{\emph{Exact vs.\ Final rule}} (trace mismatch among correct answers). 
Primary \ours (Tab.~\ref{tab:trace_diag}): windowed temporal localization 21\%$\to$0\%, revisit 40\%$\to$11\%, ordering (all option frames vs.\ the decisive option) 75\%$\to$0\%; full-video temporal localization 12\% and distance 93\% are unchanged. 

Full-scene \ours: windowed temporal localization 32\%$\to$0\%, revisit 51\%$\to$28\%, ordering 73\%$\to$0\%, full-video temporal localization 12\%, distance 93\%; paths 56\% (33/59) and routing 51\% (45/88) of correct answers have an incorrect sequence.

\textbf{\emph{Options after the trace.}} In both settings, every correct path or routing answer with an incorrect sequence has a generated sequence that is none of the options, and whenever the generated sequence is an option, \ours selects it (primary model: 22 in paths, 87 in routes; full-scene model: 33 in paths, 91 in routes). Otherwise the selected option is at minimal edit distance from the generated sequence for 61\% of paths and 69\% of the primary 0.6B \ours (full-scene 69\% and 90\%), ties included.

\textbf{\emph{Sequence errors.}} Of the 124 path reconstruction questions, the generated path differs from the reference for 105 (primary) and 97 (full-scene).
A wrong path has a \emph{different place set} when its set of legend letters differs from the reference's (primary: 67 of 105; full-scene: 66 of 98); the others contain the reference places in a different order or with different repetitions. Among routing traces with the reference's number of turns, 35\% of true right turns are predicted as left (primary).

\textbf{\emph{Patterns that do not replicate.}} For the full-scene model, accuracy is 70\% on three-visit paths ($n{=}33$) and 10\% on eight-visit paths ($n{=}10$), and description questions are lower on all seven types (49.0\% vs.\ 58.5\%); for the primary model, eight-visit paths reach 70\% and description questions are lower on four of seven types (49.0\% vs.\ 53.4\%). We therefore do not draw conclusions about path length or description grounding from these subsets.
\begin{table}[t]
\centering
\footnotesize
\setlength{\tabcolsep}{2.0pt}
\caption{\textbf{Full trace diagnostics} of primary \ours (0.6B, test). Top: intermediate predictions; bottom: facts that name the selected option, which agree with every correct answer by definition (answer-consistency checks). Rules as in Tab.~\ref{tab:trace_diag}. Distance labels each candidate near, mid, or far (see text). Agree: over all eligible questions; Mism.\,$|$\,corr.: trace mismatch among correct answers; Part.: partial agreement among correct answers (per option frame, per candidate, per turn, per stay; paths: $1-$normalized edit distance), $^\circ$ conditional on a matching number of turns or stays. $^a$The clip is a set of stage frames. $^b$Questions asking where the camera goes from the anchor.}
\label{tab:trace_diag_full}
\begin{tabular}{@{}lllrrrr@{}}
\toprule
Task & Fact & Rule & $n$ & Agree & Mism.\,$|$\,corr. & Part. \\
\midrule
Path & place seq. & exact & 124 & 15.3 & 69\% (43/62) & 72 \\
Routing & turn seq. & exact & 250 & 13.6 & 61\% (54/88) & 70$^\circ$ \\
Revisit & return visits & per visit & 133 & 43.6 & 11\% (7/65) & 71$^\circ$ \\
Loc., window & arrival frame & visit\,$\pm$1 & 189 & 89.9 & 0\% (0/157) & -- \\
Loc., video$^a$ & stage frame & exact & 173 & 57.8 & 12\% (13/111) & -- \\
Distance & labels & exact & 163 & 6.1 & 93\% (67/72) & 54 \\
\midrule
Ordering & decisive option & exact & 115 & 69.6 & 0\% (0/80) & 71 \\
Adjacency$^b$ & entered place & exact & 156 & 41.0 & 0\% (0/64) & -- \\
\bottomrule
\end{tabular}
\end{table}

\textbf{\emph{Distance labels.}} The reference trace bins the 3D distance from the anchor place to each candidate place (content positions, Sec.~\ref{sec:bench_design}) into near (${<}1.5$\,m), mid (${<}3.0$\,m), and far, then names the pick (``Nearest: C.'', or ``Nearest: C (closer than A).'' when the winner shares its bin); for a tie, whose answer is ``Equidistant'', it names both candidates (``Nearest: A and C, about equal.''). A correct distance answer does not require the entire generated label vector to match the reference, so the 93\% disagreement among correct answers measures agreement on all candidate labels rather than independent sequence recovery.

% \textbf{\emph{Partial agreement}} (primary \ours; all questions / correct answers). Ordering, per-option visit agreement 63\% / 71\% (575 / 400 options); distance, per-candidate labels 49\% / 54\%; paths, $1-$normalized edit distance 58\% / 72\%; routing, the number of turns matches in 60\% / 64\% of traces, and on those 45\% / 70\% of turns are right; revisit, the number of return stays matches in 50\% / 100\% (it equals the number of returns, i.e.\ the answer), and on those 71\% / 71\% of stays overlap the right visit.

\textbf{\emph{Evaluator audit.}} The parser extracts the facts of all 1{,}303 generated traces of the primary model. For a manual audit we drew, with seed 0, five agreeing and five disagreeing traces per row of Tab.~\ref{tab:trace_diag} (80 traces); an annotator checks whether the parsed facts match the trace text and whether the agree/disagree decision is valid, with every clip frame labelled by place and visit. The parsed facts were correct for 78 of 80 traces; both errors counted the anchor's starting stay as a revisit. Path agreement is judged on the letter sequence alone. With the original rules, the decision matched the annotator for 66 of 80 traces: 39/40 ``agree'' and 27/40 ``disagree'' decisions. Nearly all errors were over-strict disagreements:
\begin{itemize}
\item ordering, where all option frames had to match although the decisive option did (5);
\item windowed localization, where an adjacent transition frame without a visit label was equally valid (4);
\item revisit, where the starting stay was counted (2);
\item distance ties (2);
\item an adjacent full-video stage (1).
\end{itemize}
We revised the first three rules accordingly (decisive option; same visit or adjacent frame; return stays only), after which the decision matches the annotator for 76 of 80 traces.

Tab.~\ref{tab:trace_diag} uses the revised rules. Because these traces informed the revision, 76/80 is agreement after rule refinement, not an independent validation; and because the sample is stratified (five agreeing and five disagreeing traces per row), its agreement fraction is not an accuracy estimate over all test questions. 

A fresh stratified audit of the frozen rules then drew 80 other traces (seed 1; five agreeing and five disagreeing per row, none from the first audit) and was annotated in the same way. 
The parsed facts were correct for all 80 traces, and the decision was valid for 73: 39 of 40 agreements and 34 of 40 disagreements. Of the seven invalid decisions:
\begin{itemize}
\item two are distance traces whose coarse near/mid/far labels differ from the reference although the annotator found the generated judgment plausible;
\item two are full-video localization traces naming the adjacent stage frame, which the exact rule rejects;
\item one is a windowed localization trace naming the frame before the arrival, which the adjacent-frame allowance accepts but the annotator rejected;
\item one path and one revisit trace are rejected by the exact rule although the annotator accepted them, which suggests a brief visit missing from the reference.
\end{itemize}
The sample is stratified, so 73/80 describes the decisions on these strata rather than an accuracy over all test questions; most remaining disagreements are about how strict a rule should be, not about misread traces.

% \textbf{\emph{Ablations.}} Seq and Stay in Tab.~\ref{tab:abl_small} use the same parser; Stage~1 vs.\ no Stage~1 on Stay: 47 vs.\ 40 of 133 for the full-scene and 58 vs.\ 40 for the excluded-scene Stage~1. 
\textbf{\emph{Trace-first diagnosis of an external model.}} Gemini 3.1 Flash-Lite answers the 124 path and 250 routing test questions in one two-turn conversation (native API, the same split, clips, 112-px frames, low media resolution): the first turn is its main-table prompt without the options, ending with ``The answer options will be shown only after this step; do not choose an option yet'' and a request for the sequence as ``Path: P$\to$Q$\to$R'' (legend letters) or ``Turns: turn left, go straight''; the second turn adds the options with the model's reply kept in context and asks for the letter. The generated sequence is scored like \ours's trace (exact sequence; the reference equals the correct option's text for all 374 questions). All 374 replies parse.

\section{Author-led benchmark audit}
\label{sec:supp_audit}
The evaluator audit (Sec.~\ref{sec:supp_traces}) checks how generated traces are parsed and scored; this audit instead checks the benchmark's questions and references.
It was carried out by one author who took part in the place annotation.
It is therefore not independent: familiarity with the scenes and their places may raise agreement with the references, and the audit does not establish reference validity on its own.

\paragraph{Sample and procedure.}
We drew a stratified random sample of 10 questions for each of the seven task types and both video sources (140 questions; random seed 0): \bench questions from the 23 test scenes and Ego4D-derived questions from the full 955-question set (63 description- and 77 image-phrased), presented in one shuffled order.
For each question the author first saw only the inputs of the zero-shot models (question, clip frames, reference and option images, option text; frames enlarged to $224\times224$), selected an answer or ``cannot tell,'' and rated whether the frames suffice; the reference answer and trace were hidden until this answer was locked.
Every disagreement with the reference was then examined against the reference trace, the place and visit annotation of each input frame, and further frames of the walkthrough.

\paragraph{Agreement with the references.}
Under these reference-hidden conditions, the author's answers agree with the reference answer for 94.3\% of the \bench questions ($[86.2, 97.8]$, 95\% Wilson interval; 66 of 70), 97.1\% of the Ego4D-derived questions ($[90.2, 99.2]$; 68 of 70), and 95.7\% overall ($[91.0, 98.0]$; 134 of 140); every task type has at least 18 of 20 agreements, and ``cannot tell'' was never selected.
On the same 140 questions (original version), the evaluated models answer 41--51\% correctly (\ours 1.7B: 51.4\% on \bench and 41.4\% on Ego4D; Claude Opus 5.5: 47.1\% and 42.9\%).
These scores are broadly consistent with the corresponding full-set v1 accuracies.
Agreement is 130 of 132 when the author rated the frames as sufficient, 3 of 7 when only partly sufficient, and 1 of 1 when insufficient.
These figures measure agreement with the references, not independent human accuracy or reference validity.

\paragraph{The six disagreements.}
\begin{itemize}
\item \emph{Annotation error: place descriptions assigned to the wrong places} (\bench, relative distance, description phrasing). The descriptions of three places in this scene were rotated: the place shown with a whiteboard carried the description of a desk facing a window, and so on. The reference answer follows the place identities and was wrong for the question as worded; the descriptions have since been corrected (see below).
\item \emph{Insufficient visual evidence} (\bench, routing). The reference turns (two right turns of $63^\circ$ and $66^\circ$, from the camera poses) are hard to perceive from the shown frames; the author rated them as only partly sufficient and answered ``straight, straight.''
\item \emph{Visually similar places} (Ego4D, adjacency, image phrasing). Several shop aisles look alike and many clip frames between visits show no annotated place; the reference is consistent with the annotation, and the author rated the frames as only partly sufficient.
\item \emph{Author error} (Ego4D, adjacency, description phrasing). The described hallway appears in a single clip frame, directly followed by the staircase of the reference answer; the author rated the frames as only partly sufficient.
\item \emph{Author error} (\bench, path reconstruction). On re-examination the reference answer is correct.
\item \emph{Author error} (\bench, revisit counting, description phrasing). The author answered ``twice''; on re-examination the described place does not return in the clip, and the reference answer, ``never,'' is correct.
\end{itemize}

\paragraph{Description screen.}
Because a whole set of descriptions can be shifted between places without breaking any bookkeeping (each description's source keyframes belong to its place in all 1{,}268 cases), we screened every description against the frozen SigLIP embeddings of its scene's places and inspected each scene in which at least two descriptions match other places markedly better than their own.
Of the seven flagged scenes, six had descriptions assigned to the wrong places (swaps or rotations among places of one scene); the seventh is a false alarm.
We inspected every place of these six scenes, reassigned the misplaced descriptions, and rewrote those that matched no place, following the original labeling rules and collision check (20 of 1{,}268 descriptions changed).
Because the generator selects questions from place identities, not from description text, regenerating the six scenes changes only the wording of 113 description-phrased questions (40 in the test split); their answers, frames, and reference traces, and all other questions, are unchanged.
The screen detects only concentrated errors, so isolated misassigned descriptions may remain.
% The results in this paper were computed before this correction; on the 40 affected test questions, models received the earlier wording.
Results on the original wording are archived as v1.
Reported v2 results re-evaluate all models on the 40 corrected test questions, as detailed in Sec.~\ref{sec:supp_v2}.

\section{Corrected benchmark version}
\label{sec:supp_v2}
Unless explicitly marked as v1, results use the corrected benchmark version (v2). of \bench; the original version (v1), its predictions, and the checkpoint hashes are archived, and question ids are identical across versions.
v2 differs from v1 in three ways:
\begin{itemize}
    \item (i)~\emph{Place descriptions} (Sec.~\ref{sec:supp_audit}): 20 descriptions in six scenes were corrected; regenerating these scenes changes only the wording of 113 description-phrased questions (40 test, 73 training, none in validation), with answers, frames, and reference traces unchanged.
    Every reported model was re-run on the 40 test questions with the corrected wording, including refreshed SigLIP text embeddings and matching hints; predictions on all other questions are reused.
    The option-text controls were refitted on the corrected training options.
    Trained models (\ours and the fine-tuned baselines) keep their checkpoints, which were trained on the original wording of the 73 training questions; checkpoint selection is unaffected because no validation question changed.
    \item (ii)~\emph{Exclusions}: four revisit questions whose references were judged ambiguous in the review of the decisive grounding cases are excluded from every metric (1{,}401 test questions).
    \item (iii)~\emph{Visit rule}: a listed return at an unlabelled frame directly next to a frame of the reference visit counts toward that visit, since place labels leave about 30\% of revisit clip frames unlabelled; this turns two of the original 13 grounding disagreements of \ours into agreements, and the four exclusions remove four more.
\end{itemize}
Tab.~\ref{tab:supp_v1v2} compares the headline numbers; the largest Micro change is $+1.1$ points (VLM-3R 1.5B), concentrated in the reworded description questions, on which the strongest zero-shot models were near chance under the misassigned descriptions.
Detailed analyses in Secs.~\ref{sec:supp_traces} and~\ref{sec:supp_exp} that do not state otherwise were computed on v2.

\begin{table}[ht]
\centering
\footnotesize
\setlength{\tabcolsep}{3pt}
\caption{\textbf{v1 $\rightarrow$ v2} (test; Micro / Macro accuracy, \%).}
\label{tab:supp_v1v2}
\begin{tabular}{@{}lcc@{}}
\toprule
Model & Micro v1 $\rightarrow$ v2 & Macro v1 $\rightarrow$ v2 \\
\midrule
% Gemini 3.8 Flash & 52.1 $\rightarrow$ 53.3 & 55.5 $\rightarrow$ 56.7 \\   (2026-11-16: Gemini 3.8 Flash not reported)
Claude Opus 5.5 & 49.0 $\rightarrow$ 49.9 & 52.4 $\rightarrow$ 53.4 \\
GPT-6 Luna & 46.3 $\rightarrow$ 47.1 & 48.2 $\rightarrow$ 49.0 \\
Spatial-MLLM (FT) & 48.7 $\rightarrow$ 49.1 & 48.2 $\rightarrow$ 48.9 \\
VLM-3R (FT, 1.5B) & 48.0 $\rightarrow$ 49.1 & 49.1 $\rightarrow$ 50.5 \\
\ours (0.6B) & 51.3 $\rightarrow$ 51.6 & 50.7 $\rightarrow$ 51.1 \\
\ours (1.7B) & 55.8 $\rightarrow$ 56.6 & 55.3 $\rightarrow$ 56.3 \\
Text only (options) & 21.9 $\rightarrow$ 21.6 & 22.8 $\rightarrow$ 22.3 \\
\midrule
\multicolumn{3}{@{}l}{\emph{Revisit grounding of \ours (0.6B), disagree among consistent correct}} \\
& 13/69 $\rightarrow$ 7/65 & \\
\bottomrule
\end{tabular}
\end{table}

\section{The Ego4D-derived evaluation set}
\label{sec:supp_ego4d}
\paragraph{Construction.}
We use 40 Ego4D walkthroughs~\cite{grauman2022ego4d} filtered and sampled at one frame per second, so a frame index equals seconds into the video.
Places and visits are the automatic clusters of the same pipeline after annotator review, which splits and merges clusters as in \bench; review left 238 places.
Single-chunk TTT3R point maps~\cite{chen2026ttt3r} on the keyframes give each frame's camera pose and per-pixel world points, and questions use only these keyframes, so every shown frame has geometry.
Geometric validation, reliability tiers, and the view graph follow \bench, with gaze targets from the central point-map pixels instead of sensor depth; of the 238 places, 102 are Tier~A, 54 Tier~B, and 82 Tier~C.
Routing turns use camera headings projected on the plane perpendicular to the mean camera-up direction (the gravity estimate), with the \bench thresholds.
Distances are TTT3R units, converted to approximate metres where a camera-height assumption (1.55\,m) gives a plausible scale (34 of 40 videos); the six others keep their units, and their distance traces are not used.
The revisit shown-stay band is 5--10 stays (8--13 in \bench), since Ego4D videos have fewer places; Never, Once, and Twice get equal counts at every shown-stay count (24 facts each), which leaves no ``3 or more'' answers.
Place descriptions: Claude viewed two or three representative keyframes per Tier~A/B place and wrote one 6--12-word phrase, with the per-video word-overlap collision check of \bench; 132 of 156 labelled places are usable (24 flagged as mixed or told apart from another place only by movable objects).
The generator is the \bench generator with these inputs; the global answer-balancing assignments were recomputed on the 40 videos.
Result: 955 questions (temporal 508, revisit 133, paths 91, distance 72, ordering 59, routing 54, adjacency 38), 404 with description phrasing.
\paragraph{Shortcut check.}
Fitted on the \bench training scenes, the text-only model and the majority baseline reach 22.6\% and 30.6\% Micro (chance 20.7\%).
The majority baseline is above chance on temporal localization (35.2\%) because Ego4D first arrivals fall more often in the first stage (38\% of full-video and 33\% of windowed questions, vs.\ 27\% and 24\% in \bench), and on revisit counting (33.8\%, chance 25\%), where the Ego4D set has no ``3 or more'' answers.
Fitted on Ego4D itself, five folds by video, the text-only model reaches 28.9\% Micro (22.4\% Macro).
\paragraph{Evaluation.}
Every model uses its \bench settings and inputs (Tab.~\ref{tab:main}); \ours uses its Ego4D point-map features and the SigLIP inputs of the Ego4D frames. No Ego4D question, answer, or result was used for training, hyperparameter, checkpoint, or configuration selection of any model; \ours's Stage-1 pretraining used only ScanNet++ features (934 scenes, of which the primary Stage~1 excludes the 46 validation and test scenes) and no Ego4D video. The Ego4D-specific construction parameters (revisit band, answer-balancing caps, equal revisit answer counts) were set from the generated facts and answer distribution alone, before any model was evaluated on this set.
\emph{Description check.} Because Claude wrote the place descriptions, we compared description- and image-phrased accuracy: the three Claude models gain $+2.9$ to $+5.9$ points, less than the other zero-shot models ($+7.1$ to $+8.7$); this descriptive check does not exclude every form of description-generation bias. Geometric validation excludes 34\% of the Ego4D places as unreliable (Tier~C; 30\% in \bench).
Tab.~\ref{tab:ego4d_full} gives all types. Ego4D intervals use the same paired bootstrap, resampling the 40 videos (2{,}000 resamples, seed 0).
\emph{Trace diagnostics.} The \bench rules applied to the Ego4D traces, primary \ours / Qwen3-1.7B: exact agreement 8.8\% / 8.8\% on paths and 3.7\% / 3.7\% on routing; trace mismatch among correct answers 78\% / 80\% (paths), 85\% / 83\% (routing), 33\% / 33\% (revisit), 14\% / 15\% (full-video localization), and 1 of 226 / 1 of 228 (windowed localization). The \bench audit of the evaluator (Sec.~\ref{sec:supp_traces}) was not repeated on Ego4D.

\input{sec/table_ego4d_full}

% \section{Evaluation protocol details}
% \label{sec:supp_protocol}

\begin{table}[ht]
\centering
\small
\setlength{\tabcolsep}{2.4pt}
\caption{\textbf{Inputs and supervision} of the fine-tuned models. Sensor: ScanNet++ iPhone LiDAR depth and ARKit poses aligned to the laser scan; pred.: from the model's own 3D encoder on RGB. Target: answer letter, or a trace followed by the letter. Baselines use LoRA rank 64, \ours rank 16; only \ours receives the SigLIP matching hint. $^\ast$SigLIP embeddings and the matching hint from full-resolution frames.}
\label{tab:inputs}
\begin{tabular}{llll}
\toprule
Model & Geometry & Frames & Target \\
\midrule
Spatial-MLLM (FT) & pred.\ (VGGT) & native & letter \\
VLM-3R (FT) & pred.\ (CUT3R) & 112\,px & letter \\
Video-3D LLM (FT) & sensor & 112\,px & letter \\
\ours & sensor & 112\,px$^\ast$ & trace + letter \\
\bottomrule
\end{tabular}
\end{table}

\section{Additional experimental details}
\label{sec:supp_exp}
Values in this section were computed on the original version (v1, Sec.~\ref{sec:supp_v2}) unless they state otherwise; the revisit checks and the external-model results are v2.

\subsection{Setup}

In the primary setting, Stage~1 excludes all validation and test scenes; the \emph{full-scene} setting also uses their unlabeled RGB-D frames and poses, and serves as additional analysis.
Micro is accuracy over questions; Macro averages the seven task accuracies, and Traj.\ and Pair.\ average their task groups.
Differences between models come with 95\% intervals (in brackets) from a paired bootstrap over test scenes; an interval that includes zero does not establish equal performance, and scene resampling does not capture training-run variation.
\ours additionally uses reference trace supervision, sensor geometry, and a SigLIP matching hint.
Like the 3D-aware baselines but unlike the zero-shot API models, it receives walkthrough-position labels, which give each frame's relative time in the video (\eg, ``at 64\% of the video''); they bear directly on temporal localization (Sec.~\ref{sec:supp_poslabels}).
Answer accuracy therefore compares complete systems, including their inputs and inference protocols, and does not isolate any single component (inputs per model in Tab.~\ref{tab:inputs}).
Frames are $112\times112$ unless noted in Tab.~\ref{tab:main}; \ours's SigLIP inputs use full-resolution frames.

\subsection{Main results}

No evaluated model, including \ours's, exceeds 41\% on adjacency, 56\% on routing, or 51\% on revisit counting, so by answer accuracy, observed transitions, turns, and recurrence remain difficult. 
We diagnose the intermediate predictions below for \ours and, on path reconstruction, routing, and revisit counting, one external model.
We hypothesize that paths are easier for zero-shot models because legend images can be matched to clip frames one at a time.
Zero-shot, the 3D-aware Spatial-MLLM, Video-3D LLM, and the official VLM-3R (7B, on LLaVA-NeXT-Video-7B-Qwen2) reach 22.3\%, 17.1\%, and 22.9\%, below chance on revisit because they almost always answer ``3 or more times'' (VLM-3R on every revisit question); 26\% of Video-3D LLM's outputs contain no option letter, and the low accuracy may also reflect a mismatch between this input format and their pretraining.

\subsection{What answer accuracy misses}
\label{sec:supp_analysis}

\paragraph{Protocol.}
We compare generated sequences and frame-referenced visits with the reference, using exact sequence matching and visit-aware grounding (rules and evaluator check at the end of this subsection).
Path and routing sequences are generated before the options are shown, with the trace kept in context (Sec.~\ref{sec:method_finetune}); in the main-table protocol, all baselines see the options from the start.
Generated traces are evaluated as stated predictions, not as a readout of internal reasoning.
\paragraph{Multiple-choice accuracy overstates independent sequence recovery.}
Only 15\% of generated paths and 14\% of generated turn sequences match the reference, while 50\% and 35\% of the answers are correct (Tab.~\ref{tab:trace_diag}); \eg, a reference path USTPTR is generated as USTPR, omitting the return to T, or turns RLS as LLS, missing a right turn of about 100$^\circ$, and both are answered correctly.
These mismatches are often partial rather than complete failures: among correctly answered path questions, the mean normalized edit similarity to the reference is 0.72 (one minus the Levenshtein distance divided by the longer sequence's length); among correct routing answers, 64\% (56 of 88) have the reference number of turns, and within this subset turn accuracy is 70\%.

In the correct-answer cases with an incorrect generated sequence, the sequence matches none of the options; whenever it does match an option, \ours selects it (22 of 22 paths, 87 of 87 routes), and otherwise the selected option is usually close to it (at minimal edit distance for 61\% of paths and 69\% of routes).
A correct option can therefore be selected even when the preceding generated sequence is incorrect, which answer accuracy alone does not reveal.
Full-video localization disagrees for 12\% of correct answers; among the 157 correct windowed-localization answers, no mismatches were observed.
\paragraph{Correct counts can hide incorrect visit grounding.}
For revisit we separate three checks: \emph{answer accuracy} (is the predicted count correct?), \emph{trace--answer consistency} (does the number of listed returns support the predicted count?), and \emph{visit grounding} (given a consistent count, do the listed returns lie in the right visits?).
All 133 of \ours's revisit traces are count-consistent (Tab.~\ref{tab:revisit_checks}; four questions with ambiguous references excluded).
Of its 65 correct answers, 7 (11\%) list return visits (stays) that disagree with the reference visits under the visit rule: the count is right, but at least one claimed return is not in the corresponding reference visit. 

\textbf{\emph{Reference-blind review.}} Disagreement with the reference does not by itself establish a grounding error, so every decisive case was reviewed against the frames without place or visit labels: the model's and the reference's return sets were shown as X and Y in random order, without their source, and the reviewer judged whether each set's frames show the anchor place as separate returns, which set is correct, and the nature of the disagreement.

For \ours, the reviewer judges the reference correct in 6 of the 7 cases, 3 of them clear wrong-place or wrong-visit errors and 3 visit-boundary cases (the same place, but where a stay starts or ends is unclear), and both sets acceptable in 1 case (the claimed return lies two unlabelled frames from the reference visit).

The reference was never judged wrong.
Clear grounding errors therefore account for 3 of the 65 correct, count-consistent answers (5\%), and disagreements in which the reference is judged better for 6 (9\%).
For Gemini 3.1 Flash-Lite, the reference is judged correct in all 5 cases: 2 clear errors and 3 boundary cases. 

The selected cases, with the frames, the model outputs, and the classifications, are shown in Sec.~\ref{sec:supp_grounding_cases}.
The review was carried out by one author who took part in the annotation and had seen the seven \ours cases before, so it is not independent and, for \ours, not fully blind.

Fig.~\ref{fig:trace_grounding} shows such a case, in which the claimed return lies at another place and the actual return, a later visit to the anchor's corner, is missed.
Unlike sequence mismatches, these errors cannot be detected by comparing the trace with the options, since every option is only a count; they are visible only through the reference's frame--place--visit links.
\begin{table}[ht]
\centering
\footnotesize
\setlength{\tabcolsep}{3pt}
\caption{\textbf{Revisit checks} (test, corrected version, 133 questions; four with ambiguous references excluded). Consistent: the number of listed returns supports the model's own answer. Disagree: count-consistent correct answers whose listed return visits fail the visit rule (an unlabelled frame directly next to a frame of the reference visit counts toward it). Denominators: all revisit questions, correct answers, and count-consistent correct answers, respectively.}
\label{tab:revisit_checks}
\begin{tabular}{@{}lccc@{}}
\toprule
Model & \shortstack{Consistent\\(all)} & \shortstack{Inconsistent\\(correct)} & \shortstack{Disagree (consistent,\\correct)} \\
\midrule
\ours (0.6B) & 133/133 & 0/65 & 7/65 \\
Gemini 3.1 Flash-Lite & 57/133 & 26/42 & 5/16 \\
\bottomrule
\end{tabular}
\end{table}
\paragraph{An external model shows both errors.}
Under the same trace-first protocol, Gemini 3.1 Flash-Lite matches the reference for 19\% of paths and 7\% of routes, and 71\% of its correct path answers (60 of 84) and 76\% of its correct routing answers (59 of 78) come with an incorrect sequence; like \ours, it selects the option matching its sequence whenever one exists (24 of 24 paths, 58 of 58 routes).
Asked to list the anchor's return visits before answering revisit questions (options shown, as for \ours), it answers 31.6\% correctly (30.8\% without the listing), but its listed returns support its own answer for only 57 of 133 questions (Tab.~\ref{tab:revisit_checks}): most of its mismatches among correct answers are unsupported counts.
Among its 16 correct, count-consistent answers, 5 list a return visit that disagrees with the reference, so the visit-grounding error also occurs in this model, though on few count-consistent answers.
\paragraph{Sensitivity to the scoring rule.}
Both findings survive reasonable changes of the rule.
Among correct, count-consistent revisit answers, \ours (0.6B) disagrees with the reference for 7 of 65 with the adjacency tolerance, 9 of 65 without it, and 26 of 65 when exact frame ranges are required; Gemini 3.1 Flash-Lite for 5, 5, and 8 of 16.
For sequences, allowing one edit (Levenshtein distance ${\le}1$) still leaves an incorrect sequence for 50\% of \ours's correct path answers and 36\% of its correct routing answers (exact: 69\% and 61\%; mean normalized edit distance 0.32 and 0.39), and for 48\% and 34\% of Gemini 3.1 Flash-Lite's (exact: 71\% and 76\%).
\paragraph{Answer accuracy depends on the elicitation protocol.}
For Gemini, writing the sequence in a first turn and seeing the options only in a second raises answer accuracy over its single-turn, options-first score (67.7\% vs.\ 47.6\% on paths, 31.2\% vs.\ 23.6\% on routing).
Conversely, shown the options from the start, a format it was not trained on, \ours generates sequences about as accurately (13.7\% and 16.0\% agreement, vs.\ 15.3\% and 13.6\%) but answers only 21.0\% and 16.4\% correctly, because it often no longer selects the option matching its own sequence (Sec.~\ref{sec:supp_traces}); its path and routing accuracy thus depends on the trained protocol.
\paragraph{Where traces go wrong.}
Of the 105 wrong generated paths, 67 use a different set of legend places than the reference rather than only a different order, consistent with place-matching errors, omissions, or hallucinated places; on routing traces with the reference's number of turns, 35\% of true right turns (39 of 113) are predicted as left.
Questions that describe places in words are answered less often overall than image-referenced ones (49.0\% vs.\ 53.4\%), most on paths, revisit, and adjacency, but not on every type (Sec.~\ref{sec:supp_traces}).
These errors point to place grounding and turn prediction as directions for improvement; \bench supports the same diagnosis for models that produce parseable task-specific traces.
\paragraph{Rules and evaluator reliability.}
Reference traces support task-specific checks of different depth: independent sequence recovery for path reconstruction and routing, count and visit grounding for revisit, the arrival frame or stage for localization, answer-consistency checks for ordering and the evaluated adjacency subset, whose decisive option agrees with a correct answer by definition, and coarse per-candidate labels for distance (Tab.~\ref{tab:trace_diag_full}).
We parse each trace into the fact of Tab.~\ref{tab:trace_diag} and compare it with the reference trace: sequences and stage frames must match exactly;
% revisit traces must list the same number of return visits, each overlapping the corresponding reference visit (per visit); 
and an arrival frame must lie in the reference frame's visit or be the adjacent clip frame (visit\,$\pm$1).
1{,}303 of the 1{,}401 test questions have a comparable fact (the 98 adjacency questions asking where the camera \emph{never} goes do not), and the parser extracts facts from all 1{,}303 generated traces.
On a fresh stratified sample of 80 traces not used for rule development, an annotator found every extracted fact correct and the agree/disagree decision valid for 73 (39 of 40 agreements, 34 of 40 disagreements); the sample is stratified, so 73/80 is not an accuracy over all test questions (details and evaluator-development history in Sec.~\ref{sec:supp_traces}).

\subsection{Ego4D evaluation}
\label{sec:supp_ego4d_eval}

We evaluate generalization along two axes: whether our dataset construction pipeline transfers to a different source of egocentric videos, and whether \ours transfers to the resulting questions without additional training.
\paragraph{Dataset construction transfer.}
We apply the same representation with annotator-reviewed places and visits, question generator, geometric validation, shortcut controls, and trace format to 40 Ego4D walkthroughs~\cite{grauman2022ego4d}; depth and poses come from TTT3R point maps~\cite{chen2026ttt3r}, and Claude wrote the place descriptions (adaptations in Sec.~\ref{sec:supp_ego4d}).
The resulting set has 955 questions over the seven types, and option text alone does not solve it (text-only 22.6\%, fitted on Ego4D 28.9\%, training-prior majority 30.6\%, chance 20.7\%; Tab.~\ref{tab:ego4d_full}).
Places and visits were reviewed by annotators, but we have not audited the generated answers and traces against the videos independently, so we read this as evidence that the pipeline is applicable to Ego4D, not that its annotation quality is validated there.
\paragraph{Evaluation protocol.}
\ours is trained on \bench and evaluated directly on the Ego4D-derived set, without fine-tuning, hyperparameter, checkpoint, or configuration selection on that set; its Stage-1 pretraining used only ScanNet++ scenes, and it receives TTT3R geometry instead of the sensor geometry it was trained with.
All other models use their \bench settings and inputs (Tab.~\ref{tab:main}); whether external models saw Ego4D during their own pretraining cannot be established, and the Claude-written descriptions show no unusually large advantage for the Claude models (Sec.~\ref{sec:supp_ego4d}).
Temporal localization makes up 53\% of the questions (26\% in \bench), so we report Micro together with Macro and per-type accuracy; intervals resample the 40 videos.
\paragraph{Model transfer.}
Without additional training, \ours keeps its high temporal-localization accuracy (69.3\% and 71.5\%, against at most 44.3\% for the zero-shot models and 61.0\% for the 3D fine-tunes under the evaluated input protocols), while performance across task types remains mixed.
As on \bench, the advantage lies in windowed questions (0.6B: 79.3\% vs.\ 42.5\% for Gemini 3.1 Flash-Lite, the best zero-shot model on these questions); on full-video questions the difference is smaller (56.5\% vs.\ 50.2\% for Claude Opus 5.5).
On Ego4D this advantage depends on the walkthrough-position labels: without them, windowed localization falls to 40.0\%, below Gemini 3.1 Flash-Lite, and Micro to 44.1\%, while accuracy on the other six types does not drop ($+4.7$, interval includes zero).

\ours's Micro accuracy is the highest (56.0\% with Qwen3-1.7B, 54.7\% with 0.6B), but on the other six types both models are about 10 points below Claude Opus 5.5 (1.7B $-9.6$ and 0.6B $-10.1$), and their gap to the 3D fine-tunes is unresolved (1.7B vs.\ VLM-3R-1.5B $-3.6$).
Zero-shot models again lead on path reconstruction (Claude Opus 5.5 78.0\%) and adjacency (Claude Opus 5.5 50.0\%); revisit accuracy remains low for every model (29--40\%; revisit majority baseline 33.8\%, chance 25\%), and routing (54 questions) remains difficult, with zero-shot accuracies ranging from 13.0\% to 25.9\%.
Per-type results and \bench accuracy per model are in Tab.~\ref{tab:ego4d_full}.
% \paragraph{Reference traces on Ego4D.}
% Applying the same evaluator yields a similar pattern relative to the generated references: with the same rules, 78\% of primary \ours's correct path answers (28 of 36) and 85\% of its correct routing answers (11 of 13) come with an incorrect generated sequence, and 33\% of its correct revisit answers (17 of 52) fail visit grounding with count-consistent traces (19\% on \bench); windowed localization again almost never disagrees (1 of 226).
% Neither the evaluator's decisions nor the Ego4D reference traces have been independently audited against the videos, so these numbers measure consistency with the generated references.
% \paragraph{Conclusion.}
% The construction pipeline applies to Ego4D; \ours's zero-shot advantage there is confined to temporal localization and depends on walkthrough-position labels, which largely reveal localization answers (Sec.~\ref{sec:supp_poslabels}), so it is not evidence of visual transfer, and it trails the strongest zero-shot models on the other types.
% Ego4D answers, references, and evaluator decisions have not been independently validated, and one additional video source with small sets for some types (38 adjacency, 54 routing questions) does not establish general transfer.

\input{sec/supp_grounding_cases}

\clearpage
% \section{Accuracy vs.\ model size}
% \input{sec/fig_size_supp}
% Fig.~\ref{fig:size} plots test accuracy against language-model size for every model with a known size.

% 2026-11-15: moved from the main paper (5_experiments.tex Sec. 5.4) when 5_experiments_v2 dropped it
\section{Training and component analysis}
\label{sec:exp_ablation}

The clearest task-specific evidence concerns revisit counting after geometric pretraining; the overall benefits of Stage~1, stream fusion, and the primary-setting matching hint remain unresolved, and trace supervision is not isolated by the available controls.
We examine trace supervision, geometric pretraining, and the input components of \ours; primary-setting controls are reported first (they lack a geometry-only control), and full-scene analyses are labeled separately.
Most variants are evaluated with one training seed, so we read differences of a few points as indicative (per-type results in Sec.~\ref{sec:supp_ablations}).

\begin{table}[ht]
\centering
\footnotesize
\setlength{\tabcolsep}{1.2pt}
\caption{\textbf{Training and component analysis} (test; one training seed unless noted). $\Delta$: Micro difference to the first row of the block, from unrounded accuracies, with a 95\% interval from resampling the 23 test scenes. Seq: exact sequence agreement, the share of generated path and turn sequences that match the reference (pooled, $n{=}374$); Stay: visit-level stay agreement, the share of revisit traces whose listed return visits match the reference visits one by one, an unlabelled frame next to a frame of the visit counting toward it ($n{=}133$; four ambiguous questions excluded). Corrected version (v2), 1{,}401 test questions. + SigLIP: single-stream variants that keep the shared SigLIP tokens and matching hint, each starting from a Stage-1 checkpoint pretrained with the same stream. Per-type results: Sec.~\ref{sec:supp_ablations}.}
\label{tab:abl_small}
\begin{tabular}{@{}lccccc@{}}
\toprule
Variant & Macro & Micro & $\Delta$ [95\% interval] & Seq & Stay \\
\midrule
\multicolumn{6}{@{}l}{\emph{Primary setting (Stage~1 without the evaluation scenes)}} \\
\ours (primary) & 51.1 & 51.6 & -- & 14.2 & 43.6 \\
No Stage 1 & 47.7 & 50.3 & $-1.3$ $[-4.3, 1.2]$ & 15.5 & 30.1 \\
Letter only, LR $2{\times}10^{-5}$ & 23.8 & 23.2 & $-28.4$ $[-33.4, -24.5]$ & -- & -- \\
\quad second training seed & 27.8 & 28.1 & $-23.5$ $[-27.8, -19.7]$ & -- & -- \\
Letter only, cos., $10^{-4}$ & 39.0 & 39.8 & $-11.8$ $[-15.0, -8.6]$ & -- & -- \\
SigLIP tokens, no hint & 48.4 & 49.4 & $-2.2$ $[-4.9, 0.6]$ & 16.8 & 39.1 \\
No position labels & 46.9 & 45.6 & $-6.0$ $[-10.0, -2.5]$ & 15.5 & 35.3 \\
Position labels only & 27.0 & 31.5 & $-20.1$ $[-24.4, -17.0]$ & 9.9 & 19.5 \\
\midrule
\multicolumn{6}{@{}l}{\emph{Full-scene setting}} \\
\ours (full-scene) & 52.6 & 54.2 & -- & 18.7 & 35.3 \\
Appearance + SigLIP & 47.3 & 49.0 & $-5.2$ $[-9.4, -1.6]$ & 15.8 & 37.6 \\
Pose + SigLIP & 35.8 & 40.0 & $-14.3$ $[-18.0, -9.7]$ & 2.4 & 15.8 \\
Geometry + SigLIP & 53.0 & 53.3 & $-0.9$ $[-3.6, 1.9]$ & 21.9 & 39.8 \\
Letter only, LR $2{\times}10^{-4}$ & 22.8 & 21.5 & $-32.8$ $[-39.0, -27.3]$ & -- & -- \\
Letter only, LR $2{\times}10^{-5}$ & 47.1 & 45.7 & $-8.6$ $[-11.6, -5.7]$ & -- & -- \\
SigLIP tokens, no hint & 45.3 & 46.5 & $-7.8$ $[-11.7, -3.7]$ & 11.8 & 33.1 \\
No SigLIP input & 48.4 & 49.2 & $-5.1$ $[-8.8, -1.5]$ & 15.0 & 47.4 \\
\bottomrule
\end{tabular}
\end{table}

\paragraph{Trace supervision.}
Letter-only variants of \ours keep all inputs but use the answer letter as the only target, with the options shown from the start.
In the full-scene setting, the evaluated trace-supervised configuration exceeds the letter-only control by 8.6 points (54.2\% vs.\ 45.7\%, per-type changes in Sec.~\ref{sec:supp_ablations}).
This comparison includes differences in learning rate (the control uses $2{\times}10^{-5}$, the fine-tuned baselines' rate, fixed before any validation result; at the trace-supervised model's $2{\times}10^{-4}$ letter-only training collapses to 21.5\%) and in option timing, so it does not isolate trace supervision.

In the primary setting, letter-only training at $2{\times}10^{-5}$ stays near chance with two training seeds (23.2\% and 28.1\%; answer A for 94\% and 75\% of questions). We read this as an instability of letter-only training under this recipe rather than as an estimate of the benefit of traces; the letter-trained baselines were not trained with traces.
To stabilize the control, we trained it again with warm-up and cosine decay at two learning rates fixed in advance ($2{\times}10^{-5}$ and $10^{-4}$; same backbone, inputs, excluded-scene Stage~1, training questions, and 6k steps) and selected by validation accuracy: $2{\times}10^{-5}$ stays at chance (at most 22.6\% on validation), while $10^{-4}$ learns (36.4\% on validation at 6k steps, still rising; 48.9\% for trace supervision).
On the test scenes the selected control reaches 39.8\% Micro (39.0\% Macro), without a dominant answer letter (A for 23\% of questions), 11.8 points below the trace-supervised model.

\paragraph{Position labels only.}
\label{sec:supp_poslabels}
To measure what the textual inputs alone support, we trained a control that keeps \ours's recipe (trace supervision, options after the trace for path and routing, frame labels and walkthrough-position labels, 6k steps at the primary learning rate, Qwen3-0.6B with fresh LoRA) but removes every visual input: no appearance, geometry, pose, or SigLIP tokens, no matching hint, and no Stage~1. Each frame contributes only its text label (\eg, ``Context frame 3 (at 47\% of the clip)''); options that are images are reduced to their letters.

The checkpoint was selected by validation accuracy among 1.5k, 3k, 4.5k, and 6k steps (22.0, 29.5, 26.5, and 28.6\%; 3k selected), with one training seed.
On the test scenes it reaches 31.5\% Micro (27.0\% Macro), 20.1 points below \ours and 9.9 points above the text-only and majority baselines (21.6\% and 21.3\%).
Without localization questions it is at chance (20.1\% vs.\ 43.8\% for \ours on the remaining 1{,}039 questions).
The gain is concentrated in temporal localization, where it reaches 64.4\% (\ours: 74.0\%): 44.5\% on full-video questions and 82.5\% on windowed questions, where it matches \ours (83.1\%).
The cause is clip construction: a localization clip adds the arrival frame to frames sampled across the window or video, so it often lies close to another clip frame, and their position labels nearly coincide (\eg, 32\% and 33\%).

Choosing the earlier frame of the closest pair of clip frames gives the correct stage for 79.4\% of windowed questions (58.4\% of full-video questions) without looking at any image.
Localization accuracy, windowed questions above all, therefore does not show visual localization for models that receive position labels (\ours and the fine-tuned 3D baselines; the zero-shot API models receive no position labels, Tab.~\ref{tab:inputs}), and the localization trace agreement in Tab.~\ref{tab:trace_diag} should be read the same way.
The six other task types are not affected: on them the control stays at chance.
% Limitations: one training seed; the control's traces are scored with the same rules (path and turn sequences 9.9\%, visit-level stays 19.5\%), and its revisit accuracy (29.3\%) is below the text-only baseline (31.6\%).

% \paragraph{Geometric pretraining.}
% In the primary setting, removing Stage~1 changes Micro by $-1.3$ points ($[-4.3, 1.2]$), and in two other settings the gain from Stage~1 ranges from $-2.3$ to $+5.8$ points (below).
% Its largest observed effect is on revisit-count accuracy, which rises from 36.8\% to 48.9\% with the excluded-scene Stage~1.
% but the scene interval includes zero ($[-0.9, 27.8]$ over the 17 test scenes with revisit questions; $[0.8, 29.3]$ on the original version).
% Visit grounding, by contrast, is inconclusive: revisit stays match the reference visit by visit in 43.6\% vs.\ 30.1\% of revisit questions, a difference whose scene interval reaches zero ($[0.0, 28.1]$).

\paragraph{Input streams and the matching hint.}
The incremental benefit of fusing all streams is unresolved: in the full-scene setting, geometry with the shared SigLIP inputs reaches 53.3\% Micro ($-0.9$, interval includes zero) with higher Macro, Seq, and Stay than the full system.
The matching hint matters in the full-scene setting, where removing it costs 7.8 points, but in the primary setting the cost is 2.2 points with an interval that includes zero.
Per-type changes for these and the other full-scene variants are in Sec.~\ref{sec:supp_ablations}.

\section{Full ablation results}
\label{sec:supp_ablations}

\paragraph{Full-scene pretraining.}
In the full-scene setting, Stage~1 also sees the unlabeled frames, depth, and poses of the 46 validation and test scenes (no questions or labels).
With the 0.6B model it reaches 54.2\% Micro (second seed: 54.2\%), 2.6 points above the primary setting; the paired interval includes zero, so exposure to the evaluation environments does not produce a measurable gain (Tab.~\ref{tab:supp_fullscene}).
The following comparisons use v1 results; Tables~13--14 are separate from the v2 analyses in Tables~11--12.

\begin{table*}[ht]
\centering
\small
\setlength{\tabcolsep}{3.2pt}
\caption{\textbf{Full-scene Stage~1} (\bench test, corrected version, 1{,}401 questions; accuracy, \%). Columns as in Tab.~\ref{tab:main}. $^c$SigLIP embeddings from 112-px frames instead of full resolution.}
\label{tab:supp_fullscene}
\begin{tabular}{lc ccccc c cc c c c}
\toprule
Model & LLM & Ord. & Loc. & Rev. & Rte. & Pth. & Traj. & Adj. & Dst. & Pair. & Macro & Micro \\
\midrule
\ours\ (primary) & 0.6B & 71.3 & 74.0 & 48.9 & 35.2 & 50.0 & 55.9 & 33.9 & 44.2 & 39.0 & 51.1 & 51.6 \\
\midrule
\ours\ (full-scene) & 0.6B & 71.3 & 80.9 & 45.9 & 34.8 & 48.4 & 56.3 & 39.0 & 47.9 & 43.4 & 52.6 & 54.2 \\
\ours\ (full-scene), seed 2 & 0.6B & 79.1 & 77.1 & 46.6 & 36.0 & 41.9 & 56.2 & 40.2 & 50.9 & 45.5 & 53.1 & 54.2 \\
\ours\ (full-scene), 112\,px$^c$ & 0.6B & 73.0 & 80.4 & 47.4 & 30.0 & 50.0 & 56.2 & 35.8 & 41.1 & 38.5 & 51.1 & 52.3 \\
\bottomrule
\end{tabular}
\end{table*}

% Tabs.~\ref{tab:modality} and~\ref{tab:ablations} give per-type results for the analyses of Sec.~\ref{sec:exp_ablation} (Tab.~\ref{tab:abl_small}).

\textbf{\emph{Full-scene details.}} Trace vs.\ letter-only (LR $2{\times}10^{-5}$): the largest gains are on temporal localization ($+21.0$), routing ($+16.8$), and revisit counting ($+13.1$), with small losses on paths, ordering, and distance. Appearance with the shared SigLIP inputs loses most on relative distance ($-15.9$), routing ($-9.6$), and revisit ($-5.8$), and pose alone with them reaches 39.9\%. Tokens and hint together add 5.3 points over no SigLIP input (54.0\% vs.\ 48.7\%), whereas removing only the hint costs 7.7 points (46.3\%), most on localization ($-11.9$), routing ($-9.6$), and ordering ($-8.7$); tokens without the hint give no clear improvement, and the hint raises exact agreement of localization frames (62.2\% vs.\ 51.1\%) and of turn sequences (17.2\% vs.\ 9.2\%). In the primary setting, removing the hint costs 2.4 points, spread over relative distance ($-8.6$), ordering ($-7.9$), and revisit ($-5.1$), with gains on adjacency and paths.

\input{sec/table_modality}
\input{sec/table_ablations}

\paragraph{Excluded-scene Stage~1 in other settings.}
The excluded-scene Stage~1 learns its objectives on held-out scenes (validation losses keep falling: correspondence 0.30$\to$0.16, coordinate recovery 0.17$\to$0.09), which shows that the objectives are learnable, not that they help downstream.
Beyond the main configuration (Sec.~\ref{sec:exp_ablation}), we fine-tuned from the Stage~1 that excludes the 46 validation and test scenes in the 112-px setting (all inputs at 112\,px, a heading feature).
With Qwen3-0.6B it reaches 52.2\%, between the two seeds with full Stage~1 (54.8\% and 50.6\%) and 5.8 points above no Stage~1; with Qwen3-1.7B it reaches 50.7\%, against 53.0\% without Stage~1.

\section{Data, license, and ethics}
\label{sec:supp_ethics}
\bench is derived from ScanNet++~\cite{yeshwanth2023scannetpp}, which we obtained through the dataset's official access application and use under its terms of use for non-commercial research.
We release the place graphs, place descriptions, questions, traces, and the code that builds model inputs from a user's own ScanNet++ copy, and do not redistribute ScanNet++ images, depth, or poses.
ScanNet++ scenes are scans of indoor spaces captured with consent of their owners; we found no faces or personal documents in the frames used by our questions.
All manual annotation (reviewing place assignments, splitting and merging places, and checking revisits; Sec.~\ref{sec:supp_bench}) was carried out by two members of our research group as part of their research; no crowd workers or external annotators were recruited or paid.

\section{Zero-shot prompts}
\label{sec:supp_prompts}
All zero-shot models receive the same system prompt, ``You answer multiple-choice questions about an indoor egocentric video. You are shown labeled frames from one video, then a question with lettered options. Some options are themselves images. Use the frames as your only evidence.'', followed by the labeled frames, the question, the options, and the instruction ``Answer with the letter of the correct option only, in the form (X).''
API models are called by the following identifiers: \texttt{gemini-3.5-flash-lite} and \texttt{gemini-3.1-flash-lite} (Gemini API); \texttt{gpt-6-luna} and \texttt{gpt-4o} (OpenAI API); \texttt{claude-opus-5-5}, \texttt{claude-sonnet-5-5}, and \texttt{claude-haiku-5-5} (Anthropic API); and \texttt{Qwen/Qwen3-VL-235B-A22B-Instruct} (served by DeepInfra). The \bench test runs of Tab.~\ref{tab:main} were made between 2026-10-26 and 2026-11-01, the trace-first and revisit-grounding runs of Gemini 3.1 Flash-Lite on 2026-11-12 and 2026-11-14, and the Ego4D runs on 2026-11-13; providers may update the model behind an identifier, so later queries can differ.
 Gemini models use low media resolution, Claude Sonnet 5.5 runs with extended thinking off and Opus 5.5 at the lowest effort, since its thinking cannot be disabled.
Qwen3-VL-235B receives at most 32 images, so the 64- and 96-frame clips of adjacency and distance questions are thinned uniformly for it.

\section{Reproducibility}
\label{sec:supp_repro}
We will release the \bench benchmark (with its seed-0 scene split), the evaluation script, and the training code and checkpoints of \ours.

\section{Implementation details}
\label{sec:supp_impl}

\paragraph{Stage 1.}
Each objective is computed in its own forward and backward pass, and gradients are accumulated before every optimizer step.
AdamW with weight decay 0.01 and $\beta{=}(0.9,0.95)$, batch 64 with 2-step accumulation (128 clips per step), peak learning rate $2.8{\times}10^{-4}$ reached after 200 linear warm-up steps and cosine-decayed to zero over 5k steps, and gradient clipping at 1.0.
Correspondences are mutual nearest-neighbour patch pairs between two frames whose shared-frame 3D positions lie within 0.1 normalized units, using only patches with valid depth.

\paragraph{Stage 2.}
6k steps with an effective batch of 4 questions (4-step accumulation), LoRA rank 16, learning rate $2{\times}10^{-4}$ held constant after 150 warm-up steps, weight decay 0.01, gradient clipping at 1.0; validation every 1.5k steps with free-running decoding, and the best of the four validation points is kept (step 6k for both the 0.6B and 1.7B models).
Validation and test questions keep the released, letter-balanced option order.

\paragraph{Baselines.}
Fine-tuned baselines use learning rate $2\times10^{-5}$, $6{,}000$ steps with 4-step accumulation, and the checkpoint with the best accuracy on a fixed 300-question validation subset.
Video-3D LLM probes of 600 steps at both resolutions and at learning rates of $2\times10^{-5}$ and $10^{-4}$ stay at chance, and a 3{,}000-step run at $10^{-4}$ collapses to a single answer letter (21.6\%).
Zero-shot replies of Video-3D LLM that quote an option fragment instead of a letter are mapped to the option they match.
Qwen3-VL-235B receives at most 32 images, so its adjacency and distance clips are thinned.

\section{Model and training details}
\label{sec:supp_training}
This section gives the full definitions behind Sec.~\ref{sec:method_pretrain} and Sec.~\ref{sec:method_finetune}: how each input stream is turned into language-model tokens, how the SigLIP matching hint is computed, and the Stage-1 and Stage-2 objectives.
Optimizer settings are in Sec.~\ref{sec:supp_impl}.
Throughout, $D$ is the width of the language model (1{,}024 for Qwen3-0.6B, 2{,}048 for Qwen3-1.7B), $t$ indexes the frames of a clip and $k$ the $8\times8$ image patches of a frame.

\subsection{Input streams}
\label{sec:supp_streams}
\paragraph{Canonical scene frame.}
The input is $T$ frames $I_t$ with depth and camera-to-world poses $[\mathbf{R}_t|\mathbf{t}_t]$ (rotation, translation), here ScanNet++ iPhone LiDAR depth and ARKit poses aligned to the laser scan, as on RGB-D devices that track their pose.
These poses also define the place graph's geometry, but \ours never receives place identities, place positions, or the graph.
All geometry is expressed in a canonical frame anchored at the first kept camera of the scene and scaled by the root-mean-square camera distance from it:
\begin{equation}
\mathbf{x}^{\mathrm{t}}=\tfrac{1}{s}\,\mathbf{R}_1^{\top}(\mathbf{x}-\mathbf{t}_1),\qquad
s=\Big(\tfrac{1}{N}\textstyle\sum_{n=1}^{N}\|\mathbf{t}_n-\mathbf{t}_1\|^2\Big)^{1/2},
\end{equation}
where $N$ is the number of kept frames, $s$ is the scene scale (the root-mean-square distance of all the scene's camera positions from the first kept camera), $\mathbf{R}_1$ is the rotation of the first kept camera, $\mathbf{t}_1$ is the position of the first kept camera, $\mathbf{x}$ is a point in world coordinates, and $\mathbf{x}^{\mathrm{t}}$ the same point in the canonical frame; poses are transformed the same way.

\paragraph{Appearance.}
Each frame is resized to $112\times112$ and encoded by a frozen DINOv2 ViT-B/14~\cite{oquab2023dinov2} into $P{=}64$ patch features $\mathbf{f}_{t,k}\in\mathbb{R}^{768}$ (frame $t$ of the clip, patch $k=1,\dots,64$ of its $8\times8$ grid), which a linear layer maps to the model width:
\begin{equation}
\mathbf{v}_{t,k}=\mathbf{W}_{v}\mathbf{f}_{t,k}+\mathbf{e}_{v}.
\end{equation}
where $\mathbf{W}_{v}\in\mathbb{R}^{D\times768}$ is a learned linear projection and $\mathbf{e}_{v}\in\mathbb{R}^{D}$ a learned type embedding shared by all appearance tokens.

\paragraph{Geometry.}
For the same 64 patches, the unprojected depth points give an 8-d descriptor $\mathbf{d}_{t,k}=[\mathbf{x}_{t,k},c_{t,k},\mathbf{n}_{t,k},h_{t,k}]$: the mean canonical position, the fraction of valid depth pixels (confidence), the surface normal, and the height above the floor.
A two-layer MLP maps it to a token,
\begin{equation}
\mathbf{g}_{t,k}=\mathbf{W}_{g,2}\,\mathrm{GELU}(\mathbf{W}_{g,1}\mathbf{d}_{t,k})+\mathbf{e}_{g},
\end{equation}
where $\mathbf{W}_{g,1}\in\mathbb{R}^{D\times8}$ and $\mathbf{W}_{g,2}\in\mathbb{R}^{D\times D}$ are learned linear projections, and $\mathbf{e}_{g}$ is the geometry type embedding.

\paragraph{Pose.}
The canonical pose and the scene scale form $\boldsymbol{\pi}_t=[\mathrm{vec}(\mathbf{R}_t),\mathbf{t}_t,\log s]\in\mathbb{R}^{13}$, and two linear layers give two pose tokens that share an embedding of the frame's slot in the clip:
\begin{equation}
\mathbf{p}^{(i)}_t=\mathbf{W}_{p,i}\boldsymbol{\pi}_t+\mathbf{e}_{\mathrm{slot}}(t)+\mathbf{e}_{p},\qquad i\in\{1,2\}.
\end{equation}
where $\mathrm{vec}(\mathbf{R}_t)\in\mathbb{R}^{9}$ stacks the entries of the canonical rotation, $\mathbf{t}_t$ is the canonical camera position, $\mathbf{W}_{p,1},\mathbf{W}_{p,2}\in\mathbb{R}^{D\times13}$ are two learned linear projection layers (one per pose token), $\mathbf{e}_{\mathrm{slot}}(t)$ is a learned embedding of the frame's slot (its order in the clip, not its timestamp), shared by the frame's two pose tokens, and $\mathbf{e}_{p}$ is the pose type embedding.

\paragraph{Text--image matching.}
A frozen SigLIP-so400m encoder $\phi(\cdot) \in\mathbb{R}^{1152}$ embeds each full-resolution frame and each place description $q_j$; one shared linear layer maps both to tokens, with separate type embeddings:
\begin{equation}
\mathbf{u}_t=\mathbf{W}_{u}\,\phi(I_t)+\mathbf{e}_{\mathrm{img}},\qquad
\mathbf{w}_j=\mathbf{W}_{u}\,\phi(q_j)+\mathbf{e}_{\mathrm{txt}}.
\end{equation}
where $\mathbf{W}_{u}\in\mathbb{R}^{D\times1152}$ is the shared projection, and $\mathbf{e}_{\mathrm{img}},\mathbf{e}_{\mathrm{txt}}$ the type embeddings of image and text tokens.

\paragraph{Frame tokens in the prompt.}
In fine-tuning, the appearance and geometry grids are average-pooled from $8\times8$ to $4\times4$ ($\bar{\mathbf{v}}_{t,1:16}$, $\bar{\mathbf{g}}_{t,1:16}$), and each frame contributes a block of 35 tokens,
\begin{equation}
\mathbf{Z}_t=\mathrm{LN}\big([\mathbf{p}^{(1)}_t,\mathbf{p}^{(2)}_t,\bar{\mathbf{v}}_{t,1:16},\bar{\mathbf{g}}_{t,1:16},\mathbf{u}_t]\big)\cdot\gamma,
\end{equation}
where $[\cdot]$ stacks the tokens in this order, $\bar{\mathbf{v}}_{t,1:16}$ and $\bar{\mathbf{g}}_{t,1:16}$ are the 16 pooled appearance and geometry tokens, LN is a learned layer normalization applied to each token, and the learned scalar $\gamma$ is initialized to the mean norm of the language model's text embeddings divided by $\sqrt{D}$, so frame and text tokens enter at a similar scale.
The prompt is ordinary text with one \texttt{<FRAME>} marker per image; each marker is replaced by that frame's block $\mathbf{Z}_t$, and the token after each described option is replaced by its $\mathbf{w}_j$.
Pretraining, which has no text, feeds the full grids without normalization, $[\mathbf{p}^{(1)}_t,\mathbf{p}^{(2)}_t,\mathbf{v}_{t,1:64},\mathbf{g}_{t,1:64}]$ (130 tokens per frame), and no SigLIP tokens.
Separate streams let any stream be replaced by a learned placeholder and let geometric targets be read out at geometry and pose positions.

\paragraph{Language model.}
The language model (Qwen3-0.6B or Qwen3-1.7B~\cite{qwen3}) is frozen; every attention and MLP projection $\mathbf{W}$ receives a LoRA update~\cite{hu2022lora},
\begin{equation}
\mathbf{W}'=\mathbf{W}+\tfrac{\alpha}{r}\,\mathbf{B}\mathbf{A},\qquad r{=}16,\ \alpha{=}32,
\end{equation}
where $\mathbf{W}\in\mathbb{R}^{d_{\text{out}}\times d_{\text{in}}}$ is the frozen pretrained weight, $\mathbf{A}\in\mathbb{R}^{\rho\times d_{\text{in}}}$ and $\mathbf{B}\in\mathbb{R}^{d_{\text{out}}\times\rho}$ are the trainable low-rank factors of rank $\rho$, and $\alpha$ scales the update.
% Only $\mathbf{A},\mathbf{B}$, the stream projections, embeddings, and normalization are trained (DINOv2 and SigLIP stay frozen).
% and only $\mathbf{A},\mathbf{B}$, the stream projections, embeddings, and normalization are trained (DINOv2 and SigLIP stay frozen).
% The backbone weights and visual encoders remain frozen.
We train the LoRA parameters and input modules, together with the prediction heads in Stage~1

\subsection{SigLIP matching hint}
\label{sec:supp_hint}
For questions that describe places in words, a one-sentence hint states where the frozen SigLIP encoder locates each described place in the clip; it is computed from the model's inputs only (descriptions and clip frames), never from labels.
For place $j$ and clip frame $k$, the cosine similarity is standardized over the clip's frames,
\begin{equation}
z_{j,k}=\frac{S_{j,k}-\mu_j}{\sigma_j},\qquad S_{j,k}=\cos\big(\phi(q_j),\phi(I_k)\big),
\end{equation}
where $q_j$ is the description of place $j$, $I_k$ is clip frame $k$, $S_{j,k}$ their cosine similarity in the SigLIP embedding space, and $\mu_j$ and $\sigma_j$ the mean and standard deviation of $S_{j,\cdot}$ over the clip's frames, so $z_{j,m}$ measures how much better frame $m$ matches place $j$ than a typical frame of the same clip does.
% with $\mu_j$ and $\sigma_j$ the mean and standard deviation of $S_{j,\cdot}$ over the clip.
Each frame is assigned to its best-matching place if that match is strong enough,
\begin{equation}
a(k)=\arg\max_j z_{j,k}\quad\text{if}\ \max_j z_{j,k}>\tau,\ \tau{=}1.0,
\end{equation}
and otherwise to no place.
The hint lists, for each place, the clip frames assigned to it (consecutive frames merged into ranges, or ``no match''), followed by each place's two highest $z$ values, \eg, ``A text-image matcher finds the described places in the clip: start at \#1--3 and \#20, A at \#9, B no match. Match strength (z) of each place's top frames: start \#2 2.4 \#1 1.9; A \#9 1.7 \#4 0.6; B \#6 0.8 \#3 0.5.''

Places are labelled by their role in the question (start, destination, anchor, place) or by their option letter; for temporal ordering, each option is placed at the first frame it wins.

\subsection{Stage 1: geometric pretraining}
\label{sec:supp_pretrain}
Stage~1 uses clips of $T{=}8$ consecutive ScanNet++ keyframes (a frame is kept after 0.15\,m or $15^\circ$ of camera motion), with no text, QA annotations, place assignments, or visit labels.
The primary Stage~1 uses all ScanNet++ scenes except \bench's 46 validation and test scenes; the full-scene variant also uses their unlabeled frames, depth, and poses, never questions or place labels (Tab.~\ref{tab:supp_fullscene}).
VLM-3R's CUT3R encoder was also trained on ScanNet++, among other datasets~\cite{wang2025cut3r}.
Each objective reads its prediction from the last-layer hidden states $\mathbf{H}$ of the language model (with a Stage-1 LoRA adapter) through a small head, and the three losses are summed with equal weights,
\begin{equation}
\mathcal{L}_{\text{S1}}=\mathcal{L}_{\text{rec}}+\mathcal{L}_{\text{corr}}+\mathcal{L}_{\text{pose}}.
\end{equation}
where the three terms are defined below; each is computed in its own forward pass on its own input.

\paragraph{Masked geometry recovery.}
60\% of the geometry tokens of the 8-frame clip are replaced by a learned mask embedding, and the model regresses each masked patch's canonical position, weighted by depth confidence:
\begin{equation}
\mathcal{L}_{\text{rec}}=\frac{\sum_{t,k}b_{t,k}c_{t,k}\,\|\mathrm{MLP}(\mathbf{H}^{\text{geo}}_{t,k})-\mathbf{x}_{t,k}\|_1}{\sum_{t,k}b_{t,k}c_{t,k}},
\end{equation}
% where $\mu_{t,k}\in\{0,1\}$ marks masked tokens.
where $b_{t,k}\in\{0,1\}$ is 1 if geometry token $(t,k)$ is masked, $c_{t,k}$ is its depth confidence, $\mathbf{H}^{\text{geo}}_{t,k}\in\mathbb{R}^{D}$ is the last-layer hidden state at that token's position, MLP is a two-layer head ($D\to D\to3$,), and $\mathbf{x}_{t,k}$ is the patch's true canonical position; only masked patches contribute, and patches with unreliable depth count less.

\paragraph{Cross-frame correspondence.}
For two adjacent frames $i,j$, a head maps the hidden states at the geometry positions to unit-length embeddings $\mathbf{q}$, and each patch $k$ of frame $i$ must find the patch $m(k)$ of frame $j$ that shows the same surface:
\begin{equation}
\mathcal{L}_{\text{corr}}=-\frac{1}{|\mathcal{Q}|}\sum_{k\in\mathcal{Q}}\log\frac{\exp(\boldsymbol{\eta}_{i,k}^{\top}\boldsymbol{\eta}_{j,m(k)}/\tau_c)}{\sum_{l}\exp(\boldsymbol{\eta}_{i,k}^{\top}\boldsymbol{\eta}_{j,l}/\tau_c)},\qquad\tau_c{=}0.1,
\end{equation}
where $\boldsymbol{\eta}_{i,k}\in\mathbb{R}^{D/2}$ is the normalized embedding of patch $k$ of frame $i$ (a two-layer head on its geometry hidden state), $\mathcal{Q}$ is the set of patches of frame $i$ that have a correspondence (mutual nearest neighbours in canonical 3D), $m(k)$ is the corresponding patch in frame $j$, $l$ runs over all 64 patches of frame $j$, and $\tau_c$ is a temperature; the loss is a softmax classification of the correct patch among all patches, the local analogue of recognizing a revisited place.

\paragraph{Relative pose with the pose stream hidden.}
The model sees only the first and last frames of the clip, with both frames' pose tokens hidden; each frame's hidden states are averaged over its tokens, and a head on the two averages regresses the relative rotation, in the 6D representation~\cite{zhou2019continuity}, and the relative translation:
\begin{equation}
\mathcal{L}_{\text{pose}}
  =\frac{1}{6}\|\hat{\mathbf r}-\mathbf r\|_2^2
  +\frac{1}{3}\|\hat{\boldsymbol\delta}
  -\boldsymbol\delta\|_2^2.    
\end{equation}
(both averaged over their components), where $\mathbf{r}\in\mathbb{R}^{6}$ holds the first two columns of the relative rotation $\mathbf{R}_T^{\top}\mathbf{R}_1$, $\boldsymbol{\delta}=\mathbf{R}_T^{\top}(\mathbf{t}_1-\mathbf{t}_T)\in\mathbb{R}^{3}$ is the relative translation in canonical units, and $\hat{\mathbf{r}},\hat{\boldsymbol{\delta}}$ are the predictions of a two-layer head on the two frames' averaged hidden states.
The geometry tokens stay visible, so the model can infer the camera motion from appearance and canonical geometry.
Stage~1 runs for 5k steps with 128 clips per step.

\subsection{Stage 2: trace fine-tuning}
\label{sec:supp_finetune}
Stage~2 keeps the Stage-1 projections, tokenizers, and embeddings, discards the prediction heads, and adds a new LoRA adapter: keeping the Stage-1 adapter, trained without text, did not help (53.9\% vs.\ 54.8\% Micro in a full-scene variant with SigLIP at 112\,px).

\paragraph{Prompt and target.}
Each question is rendered as text with a \texttt{<FRAME>} marker after each image's label, which includes the frame's walkthrough position, \eg, ``Context frame 3 (at 64\% of the video):~\texttt{<FRAME>}''; the matching hint, when present, is appended to the question.
The target is $y=$ \texttt{Reasoning: <trace> Answer: (X)}, where the trace states the intermediate facts in terms of clip frames, \eg, ``Start at \#1--2. \#5--12: turn left. Destination at \#13--14: go straight.'' for routing (all types in Sec.~\ref{sec:supp_traces}).
With prompt $x$ (text and frame tokens), the loss is next-token cross-entropy over the trace and answer tokens,
\begin{equation}
\mathcal{L}_{\text{S2}}=-\frac{1}{|\mathcal{Y}|}\sum_{i\in\mathcal{Y}}\log p_\theta(y_i\mid x,y_{<i}),
\end{equation}
where $y_i$ is the $i$-th target token, $y_{<i}$ the target tokens before it, $p_\theta$ the language model's next-token distribution with trainable parameters $\theta$ (LoRA, projections, embeddings), and $\mathcal{Y}$ the positions of trace and answer tokens, which excludes the prompt and any options inserted after the trace.
% where $\mathcal{Y}$ excludes the prompt and any options inserted after the trace.
The trace asks the model to state place matches and visits before it answers, turning the place graph into step-level supervision; answer-only training of the same model is compared in Sec.~\ref{sec:exp_ablation}, where the comparison also differs in learning-rate schedule and option timing.

\paragraph{Options after the trace.}
For routing and path reconstruction, the prompt contains the clip, the question, and, for paths, the lettered legend images of all candidate places, but no options: the model writes its trace up to the first line break, the options are then appended to the context, and the model writes its answer with its own trace still in context, so the sequence cannot be copied from the options and the selected option can differ from it.
Training uses the same layout; for the other types the options precede the trace.
Options are randomly permuted whenever a training question is loaded.

\paragraph{Decoding.}
At test time the model decodes greedily and writes its own trace; the last parenthesized letter of its output is the answer.
We fine-tune for 6k steps (effective batch 4, learning rate $2\times10^{-4}$) and keep the checkpoint with the best validation accuracy.

%% file: sec/table_ego4d_full.tex
% generated by figures/make_ego4d_tables.py from NaviMLLM/baseline_runs/ego4d_* -- do not edit by hand
\begin{table*}[t]
\centering
\small
\setlength{\tabcolsep}{3.0pt}
\caption{\textbf{Ego4D-derived evaluation set, results by question type} (accuracy, \%; 955 questions: ordering 59, localization 508, revisit 133, routing 54, paths 91, adjacency 38, distance 72; chance 20\%, 25\% for revisit). SN++: Macro on the \bench test scenes. $^a$Fitted on the \bench training scenes. $^\dagger$Frames at 512\,px. Bold / underline: best / second (lower bounds not ranked).}
\label{tab:ego4d_full}
\begin{tabular}{@{}llcccccccccc@{}}
\toprule
Model & LLM & Ord. & Loc. & Rev. & Rte. & Pth. & Adj. & Dst. & Macro & Micro & SN++ \\
\midrule
Chance & -- & 20.0 & 20.0 & 25.0 & 20.0 & 20.0 & 20.0 & 20.0 & 20.7 & 20.7 & 20.7 \\
Majority (training prior)$^a$ & -- & 20.3 & 35.2 & 33.8 & 20.4 & 22.0 & 23.7 & 22.2 & 25.4 & 30.6 & 21.2 \\
Text only (options)$^a$ & -- & 20.3 & 19.9 & 33.8 & 22.2 & 26.4 & 15.8 & 22.2 & 23.0 & 22.6 & 22.3 \\
\midrule
% Gemini 3.8 Flash & -- & \textbf{67.8} & 49.0 & \textbf{39.8} & 14.8 & \underline{76.9} & \textbf{63.2} & 37.5 & \textbf{49.9} & 49.3 & 55.5 \\
Claude Opus 5.5 (low effort) & -- & \underline{66.1} & 44.3 & 37.6 & 20.4 & \textbf{78.0} & \underline{50.0} & 34.7 & \underline{47.3} & 46.1 & 53.4 \\
Claude Sonnet 5.5 (no thinking) & -- & 59.3 & 38.0 & 37.6 & 16.7 & \underline{76.9} & 39.5 & 22.2 & 41.5 & 40.6 & 49.0 \\
Claude Haiku 5.5 & -- & 55.9 & 40.2 & 35.3 & 14.8 & 68.1 & 47.4 & 33.3 & 42.2 & 41.5 & 44.8 \\
Gemini 3.1 Flash-Lite & -- & 59.3 & 43.7 & 34.6 & 13.0 & 51.6 & 42.1 & 26.4 & 38.7 & 41.0 & 39.2 \\
GPT-4o & -- & 44.1 & 34.8 & 38.3 & 20.4 & 47.3 & 21.1 & 23.6 & 32.8 & 34.9 & 34.0 \\
Qwen2.5-VL$^\dagger$ & 7B & 39.0 & 25.0 & 29.3 & 14.8 & 40.7 & 26.3 & 30.6 & 29.4 & 27.9 & 26.9 \\
Qwen3-VL$^\dagger$ & 235B & 52.5 & 40.7 & 27.1 & 25.9 & 61.5 & 34.2 & 27.8 & 38.5 & 39.5 & 39.7 \\
\midrule
Spatial-MLLM (FT) & 3B & 47.5 & 61.0 & 30.8 & \textbf{33.3} & 51.6 & 26.3 & 23.6 & 39.2 & 49.3 & 48.9 \\
VLM-3R (FT) & 0.5B & 35.6 & 30.5 & \underline{39.1} & \underline{31.5} & 34.1 & 15.8 & 22.2 & 29.8 & 31.2 & 31.7 \\
VLM-3R (FT) & 1.5B & 62.7 & 51.6 & 38.3 & \textbf{33.3} & 59.3 & 15.8 & 30.6 & 41.7 & 47.1 & 50.5 \\
\midrule
\ours, Qwen3-0.6B & 0.6B & 52.5 & \underline{69.3} & \underline{39.1} & 24.1 & 39.6 & 26.3 & \underline{38.9} & 41.4 & \underline{54.7} & 51.1 \\
\ours, Qwen3-1.7B & 1.7B & 57.6 & \textbf{71.5} & 34.6 & 22.2 & 44.0 & 23.7 & \textbf{43.1} & 42.4 & \textbf{56.0} & 56.3 \\
\bottomrule
\end{tabular}
\end{table*}

%% file: sec/supp_grounding_cases.tex
% generated by analysis/make_grounding_case_appendix.py -- do not edit by hand
\section{Decisive visit-grounding cases}
\label{sec:supp_grounding_cases}
\begin{figure*}[p]
\centering
\includegraphics[width=\textwidth]{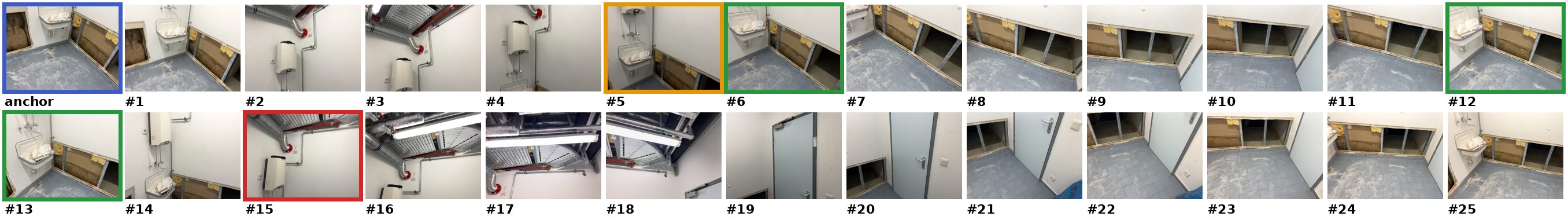}
\caption{\textbf{\ours (0.6B), case 1} (\texttt{scannetpp\_77b40ce601\_revisit\_003}). After this point, how many more times does the camera return to the location pictured in the anchor frame?
Model output: \texttt{Reasoning: Anchor room: \#1 (start), \#5, \#15. Returns: 2. Answer: (A)}. Return visits: model \#5, \#15; reference \#5--6, \#12--13.
Review: reference set judged correct; \emph{clear grounding error}. The model's return at \#15 shows the water heater on another wall, not the sink corner of the anchor; the reference returns (\#5--6, \#12--13) show the sink corner.}
\label{fig:gcase_tracer06b_2}
\end{figure*}

\begin{figure*}[p]
\centering
\includegraphics[width=\textwidth]{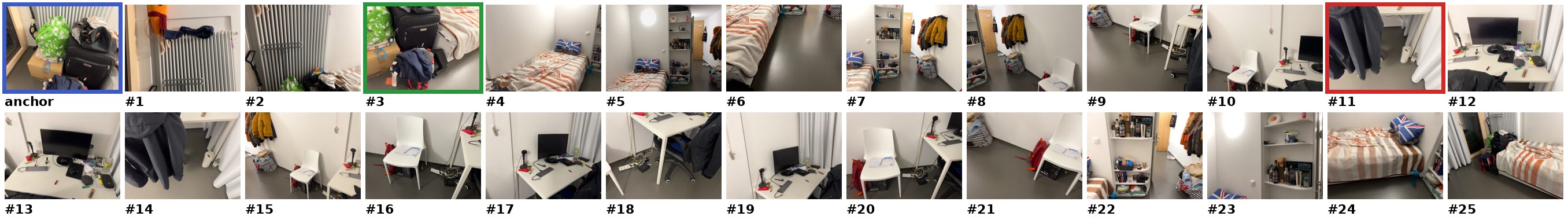}
\caption{\textbf{\ours (0.6B), case 2} (\texttt{scannetpp\_98cc0b7c9a\_revisit\_011}). After this point, how many more times does the camera return to the location pictured in the anchor frame?
Model output: \texttt{Reasoning: Anchor room: \#1-2 (start), \#11. Returns: 1. Answer: (A)}. Return visits: model \#11; reference \#3.
Review: reference set judged correct; \emph{clear grounding error}. The model's return at \#11 shows the jacket and curtain corner; the reference return \#3 shows the bag pile of the anchor.}
\label{fig:gcase_tracer06b_5}
\end{figure*}

\begin{figure*}[p]
\centering
\includegraphics[width=\textwidth]{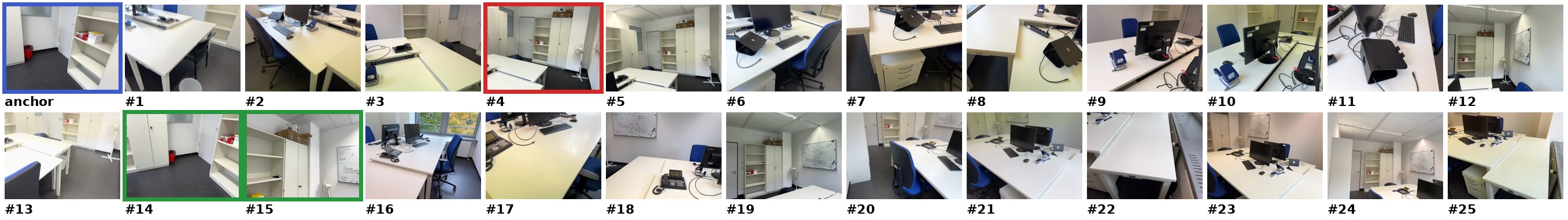}
\caption{\textbf{\ours (0.6B), case 3} (\texttt{scannetpp\_f8d5147d1d\_revisit\_002}). How many more times does the camera come back to red wastebasket beside a white door and open shelving with boxes later in this clip?
Model output: \texttt{Reasoning: Anchor room: \#4. Returns: 1. Answer: (A)}. Return visits: model \#4; reference \#14--15.
Review: reference set judged correct; \emph{visit-boundary disagreement}. The model's return at \#4 shows the shelving from across the room; the reference return \#14--15 shows the red bin and shelving up close.}
\label{fig:gcase_tracer06b_7}
\end{figure*}

\begin{figure*}[p]
\centering
\includegraphics[width=\textwidth]{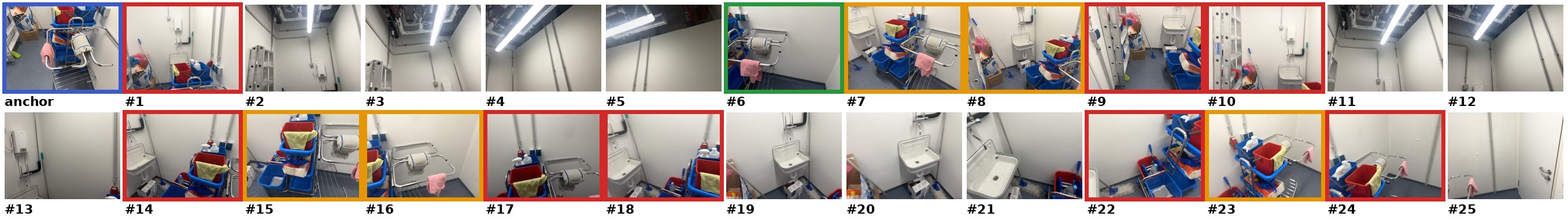}
\caption{\textbf{Gemini 3.1 Flash-Lite, case 1} (\texttt{scannetpp\_0a5c013435\_revisit\_011}). After this point, how many more times does the camera return to the location pictured in the anchor frame?
Model output: \texttt{Anchor room: \#1, \#7-10, \#14-18, \#22-24 Returns: 3 (A)}. Return visits: model \#1, \#7--10, \#14--18, \#22--24; reference \#6--8, \#15--16, \#23.
Review: reference set judged correct; \emph{visit-boundary disagreement}. The cart stays in view across neighbouring frames of this small room; the model splits the clip into four stays, the reference into three (same count bin).}
\label{fig:gcase_gemini31fl_2}
\end{figure*}

Figs.~\ref{fig:gcase_tracer06b_2}--\ref{fig:gcase_gemini31fl_2} show some correct, count-consistent revisit answer whose listed return visits disagree with the reference.
Each shows the anchor image and the clip frames in temporal order; borders mark the return visits listed by the reference (green), by the model (red), or by both (orange).
The classification is from the review of Sec.~\ref{sec:supp_analysis}, carried out reference-blind (the two sets shown in random order, without their source) by one author who took part in the annotation and had seen the seven \ours cases before; it is not independent validation.

% \begin{figure*}[p]
% \centering
% \includegraphics[width=\textwidth]{grounding_cases/tracer06b_1}
% \caption{\textbf{\ours (0.6B), case 1} (\texttt{scannetpp\_1fb6538f81\_revisit\_011}). How many more times does the camera come back to the area shown in the anchor frame later in this clip?
% Model output: \texttt{Reasoning: Anchor room: \#7. Returns: 1. Answer: (A)}. Return visits: model \#7; reference \#4.
% Review: both sets judged acceptable; \emph{visit-boundary disagreement}. \#4--\#7 are one continuous approach to the dining table; the model's \#7 and the reference's \#4 lie two unlabelled frames apart.}
% \label{fig:gcase_tracer06b_1}
% \end{figure*}

% \begin{figure*}[p]
% \centering
% \includegraphics[width=\textwidth]{grounding_cases/tracer06b_3}
% \caption{\textbf{\ours (0.6B), case 3} (\texttt{scannetpp\_77b40ce601\_revisit\_004}). How many more times does the camera come back to yellow fiberglass insulation exposed in a wall stud cavity beside a plumbing sink later in this clip?
% Model output: \texttt{Reasoning: Anchor room: \#1 (start), \#5, \#15. Returns: 2. Answer: (B)}. Return visits: model \#5, \#15; reference \#5--6, \#12--13.
% Review: reference set judged correct; \emph{clear grounding error}. Same clip as the previous case (description phrasing): the model's return at \#15 shows the water heater, not the described sink corner.}
% \label{fig:gcase_tracer06b_3}
% \end{figure*}

% \begin{figure*}[p]
% \centering
% \includegraphics[width=\textwidth]{grounding_cases/tracer06b_4}
% \caption{\textbf{\ours (0.6B), case 4} (\texttt{scannetpp\_98cc0b7c9a\_revisit\_001}). How many more times does the camera come back to the room shown in the anchor frame later in this clip?
% Model output: \texttt{Reasoning: Anchor room: \#2-4, \#6. Returns: 2. Answer: (B)}. Return visits: model \#2--4, \#6; reference \#5--6, \#20.
% Review: reference set judged correct; \emph{visit-boundary disagreement}. The two sets differ in where the returns start and end (model \#2--4 and \#6; reference \#5--6 and \#20).}
% \label{fig:gcase_tracer06b_4}
% \end{figure*}

% \begin{figure*}[p]
% \centering
% \includegraphics[width=\textwidth]{grounding_cases/tracer06b_6}
% \caption{\textbf{\ours (0.6B), case 6} (\texttt{scannetpp\_98cc0b7c9a\_revisit\_012}). How many more times does the camera come back to green printed bag and black rolling suitcase stacked by radiator later in this clip?
% Model output: \texttt{Reasoning: Anchor room: \#1-2 (start), \#8. Returns: 1. Answer: (A)}. Return visits: model \#8; reference \#3.
% Review: reference set judged correct; \emph{visit-boundary disagreement}. The model treats \#1--2 (the radiator beside the bags) as the starting stay and places the return at \#8; the reference counts \#3 as the return. Whether the radiator view belongs to the anchor's stay is unclear.}
% \label{fig:gcase_tracer06b_6}
% \end{figure*}

% \begin{figure*}[p]
% \centering
% \includegraphics[width=\textwidth]{grounding_cases/gemini31fl_1}
% \caption{\textbf{Gemini 3.1 Flash-Lite, case 1} (\texttt{scannetpp\_0a5c013435\_revisit\_006}). How many more times does the camera come back to the cleaning cart beside a second cart with an orange-wrapped pack and blue dustpan later in this clip?
% Model output: \texttt{Anchor room: \#5-6, \#9-10, \#12, \#15-16, \#19-20 Returns: 4 (A)}. Return visits: model \#5--6, \#9--10, \#12, \#15--16, \#19--20; reference \#5, \#11, \#15, \#19--20.
% Review: reference set judged correct; \emph{clear grounding error}. The model lists returns at \#9--10 and \#12, around the sink, where the reference has a single return at \#11.}
% \label{fig:gcase_gemini31fl_1}
% \end{figure*}

%% file: sec/table_modality.tex
% copied from ../experiments.tex (generated by make_results_tables.py, 2026-11-06); re-copy after regenerating
% (generated by make_results_tables.py -- do not edit by hand)
\begin{table*}[t]
\centering
\small
\setlength{\tabcolsep}{3.2pt}
\caption{\textbf{Input-stream ablation} of \ours\ on the \bench\ test split (v1, 1,405 questions).
Fine-tuning keeps only the named stream (DINOv2 appearance tokens, camera-pose tokens, or per-patch geometry tokens); the SigLIP image tokens and text-matcher hint stay in every variant.
Full S1: fine-tuned from the all-streams Stage-1 checkpoint; Matched: from a Stage-1 checkpoint pretrained with the same single stream.
Geometry comes from iPhone LiDAR depth and mesh-aligned ARKit poses.}
\label{tab:modality}
\begin{tabular}{llc ccccc c cc c c c}
\toprule
& Model & LLM & Ord. & Loc. & Rev. & Rte. & Pth. & Traj. & Adj. & Dst. & Pair. & All & Micro \\
\midrule
\multirow{4}{*}{\rotatebox{90}{\scriptsize Full S1}} & All streams (\ours) & 0.6B & 67.8 & 80.7 & 47.4 & 35.2 & 47.6 & 55.7 & 39.4 & 46.6 & 43.0 & 52.1 & 54.0 \\
 & Appearance only & 0.6B & 62.6 & 75.7 & 35.8 & 19.2 & 43.5 & 47.4 & 31.1 & 25.8 & 28.4 & 42.0 & 44.0 \\
 & Pose only & 0.6B & 68.7 & 76.2 & 37.2 & 17.2 & 29.8 & 45.8 & 28.0 & 23.3 & 25.6 & 40.1 & 42.3 \\
 & Geometry only & 0.6B & 60.9 & 78.5 & 39.4 & 27.2 & 52.4 & 51.7 & 29.9 & 56.4 & 43.2 & 49.2 & 50.5 \\
\midrule
\multirow{3}{*}{\rotatebox{90}{\scriptsize Matched}} & Appearance only & 0.6B & 68.7 & 78.5 & 41.6 & 25.6 & 45.2 & 51.9 & 36.2 & 30.7 & 33.4 & 46.6 & 48.5 \\
 & Pose only & 0.6B & 40.0 & 76.2 & 29.9 & 18.0 & 28.2 & 38.5 & 26.8 & 30.1 & 28.4 & 35.6 & 39.9 \\
 & Geometry only & 0.6B & 73.0 & 74.9 & 52.6 & 40.0 & 47.6 & 57.6 & 29.9 & 47.9 & 38.9 & 52.3 & 52.7 \\
\bottomrule
\end{tabular}
\end{table*}

%% file: sec/table_ablations.tex
% copied from ../experiments.tex (generated by make_results_tables.py, 2026-11-07); re-copy after regenerating
% (generated by make_results_tables.py -- do not edit by hand)
\begin{table*}[t]
\centering
\small
\setlength{\tabcolsep}{3.0pt}
\caption{\textbf{Ablations} of \ours\ on the \bench\ test split (v1, 1,405 questions; single runs; two seeds of \ours\ differ by 0.5 points overall).
LLM: Qwen3-0.6B (0.6B), Qwen3-1.7B (1.7B), or Qwen2.5-0.5B (0.5B).
\emph{Pretrain}: the Stage-1 checkpoint the fine-tune starts from; $^a$a second Stage-1 run whose held-out losses rose during pretraining.
\emph{SigLIP}: with or without the frozen SigLIP image and option tokens and the matching hint, for two Qwen2.5-0.5B models fine-tuned from different Stage-1 checkpoints (S1-B, S1-C);
$^b$\ours\ with its SigLIP embeddings computed from the 112$\times$112 frames instead of the full-resolution frames.
\emph{Scene memory}: $^c$the point-cloud encoder of the original design, trained from scratch during fine-tuning;
$^d$a whole-scene version with 32 or 128 tokens, added through a zero-initialized gate while \ours\ is trained for 3k more steps, compared with the same continuation without it (control).
\emph{LLM}: $^e$fine-tuned without Stage~1 (Stage-1 tokenizers are sized for the 0.6B model), with SigLIP at 112\,px and a gravity-aligned heading feature; 1.7B vs.\ 0.6B:
$^f$\ours\ fine-tuned with the answer letter as its only target.}
\label{tab:ablations}
\begin{tabular}{llc ccccc c cc c c c}
\toprule
& Model & LLM & Ord. & Loc. & Rev. & Rte. & Pth. & Traj. & Adj. & Dst. & Pair. & All & Micro \\
\midrule
\multirow{3}{*}{\rotatebox{90}{\scriptsize Pretrain}} & Stage 1 (\ours) & 0.6B & 67.8 & 80.7 & 47.4 & 35.2 & 47.6 & 55.7 & 39.4 & 46.6 & 43.0 & 52.1 & 54.0 \\
 & Stage 1, rising loss$^a$ & 0.6B & 78.3 & 71.8 & 36.5 & 28.8 & 46.0 & 52.3 & 34.3 & 42.3 & 38.3 & 48.3 & 48.8 \\
 & No Stage 1 & 0.6B & 64.3 & 75.4 & 36.5 & 36.8 & 36.3 & 49.9 & 37.8 & 44.2 & 41.0 & 47.3 & 50.0  \\
\midrule
 & No traces (letter only)$^f$ & 0.6B & 21.7 & 19.9 & 36.5 & 19.6 & 27.4 & 25.0 & 15.7 & 19.6 & 17.7 & 22.9 & 21.5  \\
\midrule
\multirow{6}{*}{\rotatebox{90}{\scriptsize SigLIP}} & S1-B, with SigLIP & 0.5B & 67.8 & 78.2 & 34.3 & 27.2 & 48.4 & 51.2 & 28.7 & 38.0 & 33.4 & 46.1 & 47.8 \\
 & S1-B, w/o SigLIP & 0.5B & 53.9 & 71.0 & 40.9 & 23.6 & 36.3 & 45.1 & 26.0 & 28.8 & 27.4 & 40.1 & 42.1 \\
 & S1-C, with SigLIP & 0.5B & 69.6 & 70.4 & 33.6 & 27.2 & 44.4 & 49.0 & 31.5 & 29.4 & 30.5 & 43.7 & 45.0 \\
 & S1-C, w/o SigLIP & 0.5B & 37.4 & 66.0 & 37.2 & 28.0 & 36.3 & 41.0 & 28.0 & 30.7 & 29.3 & 37.7 & 40.5 \\
 & No S1, w/o SigLIP & 0.5B & 54.8 & 64.6 & 34.3 & 24.8 & 39.5 & 43.6 & 26.0 & 26.4 & 26.2 & 38.6 & 40.1 \\
 & \ours, SigLIP at 112\,px$^b$ & 0.6B & 71.3 & 79.8 & 46.7 & 30.0 & 49.2 & 55.4 & 35.4 & 41.1 & 38.3 & 50.5 & 51.8 \\
\midrule
\multirow{4}{*}{\rotatebox{90}{\scriptsize Scene mem.}} & + scene memory$^c$ & 0.6B & 67.8 & 75.7 & 43.8 & 31.2 & 50.8 & 53.9 & 30.7 & 45.4 & 38.1 & 49.3 & 50.2 \\
 & +3k steps (control) & 0.6B & 73.0 & 77.3 & 42.3 & 36.8 & 61.3 & 58.2 & 39.0 & 50.3 & 44.6 & 54.3 & 54.9 \\
 & +3k steps, gated, 32 tok.$^d$ & 0.6B & 68.7 & 75.7 & 37.2 & 33.6 & 54.0 & 53.8 & 40.9 & 54.6 & 47.8 & 52.1 & 53.2 \\
 & +3k steps, gated, 128 tok.$^d$ & 0.6B & 71.3 & 80.4 & 44.5 & 35.6 & 51.6 & 56.7 & 42.5 & 52.1 & 47.3 & 54.0 & 55.5 \\
\midrule
\multirow{2}{*}{\rotatebox{90}{\scriptsize LLM}} & Qwen3-0.6B, no Stage 1$^e$ & 0.6B & 69.6 & 63.8 & 39.4 & 28.0 & 46.0 & 49.4 & 36.2 & 37.4 & 36.8 & 45.8 & 45.9 \\
 & Qwen3-1.7B, no Stage 1$^e$ & 1.7B & 67.8 & 78.7 & 49.6 & 34.4 & 50.0 & 56.1 & 35.4 & 41.1 & 38.3 & 51.0 & 52.4 \\
\bottomrule
\end{tabular}
\end{table*}

%% file: main.bib
@misc{anthropic2025claudeopus41,
  author = {{Anthropic}},
  title  = {{Claude Opus 4.1}},
  year   = {2025},
  month  = aug,
  url    = {https://www.anthropic.com/news/claude-opus-4-1}
}

@inproceedings{yang2025thinking,
  title     = {Thinking in Space: How Multimodal Large Language Models See, Remember, and Recall Spaces},
  author    = {Yang, Jihan and Yang, Shusheng and Gupta, Anjali W. and Han, Rilyn and Fei-Fei, Li and Xie, Saining},
  booktitle = {CVPR},
  year      = {2025}
}

@inproceedings{yeshwanth2023scannetpp,
  title     = {{ScanNet++}: A High-Fidelity Dataset of {3D} Indoor Scenes},
  author    = {Yeshwanth, Chandan and Liu, Yueh-Cheng and Nie{\ss}ner, Matthias and Dai, Angela},
  booktitle = {ICCV},
  year      = {2023}
}

@inproceedings{azuma2022scanqa,
  title     = {{ScanQA}: {3D} Question Answering for Spatial Scene Understanding},
  author    = {Azuma, Daichi and Miyanishi, Taiki and Kurita, Shuhei and Kawanabe, Motoaki},
  booktitle = {CVPR},
  year      = {2022}
}

@inproceedings{ma2023sqa3d,
  title     = {{SQA3D}: Situated Question Answering in {3D} Scenes},
  author    = {Ma, Xiaojian and Yong, Silong and Zheng, Zilong and Li, Qing and Liang, Yitao and Zhu, Song-Chun and Huang, Siyuan},
  booktitle = {ICLR},
  year      = {2023}
}

@inproceedings{chen2024spatialvlm,
  title     = {{SpatialVLM}: Endowing Vision-Language Models with Spatial Reasoning Capabilities},
  author    = {Chen, Boyuan and others},
  booktitle = {CVPR},
  year      = {2024}
}

@inproceedings{cheng2024spatialrgpt,
  title     = {{SpatialRGPT}: Grounded Spatial Reasoning in Vision-Language Models},
  author    = {Cheng, An-Chieh and others},
  booktitle = {NeurIPS},
  year      = {2024}
}

@article{wu2025spatialmllm,
  title   = {{Spatial-MLLM}: Boosting {MLLM} Capabilities in Visual-based Spatial Intelligence},
  author  = {Wu, Diankun and others},
  journal = {arXiv preprint arXiv:2505.23747},
  year    = {2025}
}

@inproceedings{grauman2022ego4d,
  title     = {{Ego4D}: Around the World in 3,000 Hours of Egocentric Video},
  author    = {Grauman, Kristen and others},
  booktitle = {CVPR},
  year      = {2022}
}

@inproceedings{majumdar2024openeqa,
  title     = {{OpenEQA}: Embodied Question Answering in the Era of Foundation Models},
  author    = {Majumdar, Arjun and others},
  booktitle = {CVPR},
  year      = {2024}
}

@inproceedings{chandrasegaran2024hourvideo,
  title     = {{HourVideo}: 1-Hour Video-Language Understanding},
  author    = {Chandrasegaran, Keshigeyan and others},
  booktitle = {NeurIPS Datasets and Benchmarks},
  year      = {2024}
}

@inproceedings{nagarajan2020egotopo,
  title     = {{Ego-Topo}: Environment Affordances from Egocentric Video},
  author    = {Nagarajan, Tushar and Li, Yanghao and Feichtenhofer, Christoph and Grauman, Kristen},
  booktitle = {CVPR},
  year      = {2020}
}

@inproceedings{hong20233dllm,
  title     = {{3D-LLM}: Injecting the {3D} World into Large Language Models},
  author    = {Hong, Yining and others},
  booktitle = {NeurIPS},
  year      = {2023}
}

@inproceedings{huang2024leo,
  title     = {An Embodied Generalist Agent in {3D} World},
  author    = {Huang, Jiangyong and others},
  booktitle = {ICML},
  year      = {2024}
}

@inproceedings{chen2024ll3da,
  title     = {{LL3DA}: Visual Interactive Instruction Tuning for Omni-{3D} Understanding, Reasoning, and Planning},
  author    = {Chen, Sijin and others},
  booktitle = {CVPR},
  year      = {2024}
}

@inproceedings{huang2024chatscene,
  title     = {{Chat-Scene}: Bridging {3D} Scene and Large Language Models with Object Identifiers},
  author    = {Huang, Haifeng and others},
  booktitle = {NeurIPS},
  year      = {2024}
}

@article{zhu2024llava,
  title={Llava-3d: A simple yet effective pathway to empowering lmms with 3d-awareness},
  author={Zhu, Chenming and Wang, Tai and Zhang, Wenwei and Pang, Jiangmiao and Liu, Xihui},
  journal={arXiv preprint arXiv:2409.18125},
  year={2024}
}

@inproceedings{zheng2025video3dllm,
  title     = {{Video-3D LLM}: Learning Position-Aware Video Representation for {3D} Scene Understanding},
  author    = {Zheng, Duo and Huang, Shijia and Wang, Liwei},
  booktitle = {CVPR},
  year      = {2025}
}

@inproceedings{qi2026gpt4scene,
  title={Gpt4scene: Understand 3d scenes from videos with vision-language models},
  author={Qi, Zhangyang and Zhang, Zhixiong and Fang, Ye and Wang, Jiaqi and Zhao, Hengshuang},
  booktitle={International Conference on Learning Representations},
  volume={2026},
  pages={157277--157302},
  year={2026}
}

@inproceedings{wang2025cut3r,
  title     = {Continuous {3D} Perception Model with Persistent State},
  author    = {Wang, Qianqian and Zhang, Yifei and Holynski, Aleksander and Efros, Alexei A. and Kanazawa, Angjoo},
  booktitle = {CVPR},
  year      = {2025}
}

@article{zheng2026learning,
  title={Learning from videos for 3d world: Enhancing mllms with 3d vision geometry priors},
  author={Zheng, Duo and Li, Yanyang and Wang, Liwei and others},
  journal={Advances in neural information processing systems},
  volume={38},
  pages={20560--20586},
  year={2026}
}

@inproceedings{fan2026vlm,
  title={Vlm-3r: Vision-language models augmented with instruction-aligned 3d reconstruction},
  author={Fan, Zhiwen and Zhang, Jian and Li, Renjie and Zhang, Junge and Chen, Runjin and Hu, Hezhen and Wang, Kevin and Wang, Peihao and Qu, Huaizhi and Zhou, Shijie and others},
  booktitle={Proceedings of the IEEE/CVF conference on computer vision and pattern recognition},
  pages={31054--31065},
  year={2026}
}

@inproceedings{wang2025ross3d,
  title={Ross3d: Reconstructive visual instruction tuning with 3d-awareness},
  author={Wang, Haochen and Zhao, Yucheng and Wang, Tiancai and Fan, Haoqiang and Zhang, Xiangyu and Zhang, Zhaoxiang},
  booktitle={2025 IEEE/CVF International Conference on Computer Vision (ICCV)},
  pages={9275--9286},
  year={2025},
  organization={IEEE}
}

@article{huang20263drs,
  title={3drs: Mllms need 3d-aware representation supervision for scene understanding},
  author={Huang, Xiaohu and Wu, Jingjing and Xie, Qunyi and Han, Kai},
  journal={Advances in Neural Information Processing Systems},
  volume={38},
  pages={67961--67988},
  year={2026}
}

@inproceedings{arandjelovic2016netvlad,
  title     = {{NetVLAD}: {CNN} Architecture for Weakly Supervised Place Recognition},
  author    = {Arandjelovi{\'c}, Relja and Gronat, Petr and Torii, Akihiko and Pajdla, Tomas and Sivic, Josef},
  booktitle = {CVPR},
  year      = {2016}
}

@article{keetha2023anyloc,
  title={Anyloc: Towards universal visual place recognition},
  author={Keetha, Nikhil and Mishra, Avneesh and Karhade, Jay and Jatavallabhula, Krishna Murthy and Scherer, Sebastian and Krishna, Madhava and Garg, Sourav},
  journal={IEEE Robotics and Automation Letters},
  volume={9},
  number={2},
  pages={1286--1293},
  year={2023},
  publisher={IEEE}
}

@article{murartal2015orbslam,
  title   = {{ORB-SLAM}: A Versatile and Accurate Monocular {SLAM} System},
  author  = {Mur-Artal, Ra{\'u}l and Montiel, J. M. M. and Tard{\'o}s, Juan D.},
  journal = {IEEE Transactions on Robotics},
  year    = {2015}
}

@inproceedings{savinov2018sptm,
  title     = {Semi-Parametric Topological Memory for Navigation},
  author    = {Savinov, Nikolay and Dosovitskiy, Alexey and Koltun, Vladlen},
  booktitle = {ICLR},
  year      = {2018}
}

@inproceedings{chaplot2020nts,
  title     = {Neural Topological {SLAM} for Visual Navigation},
  author    = {Chaplot, Devendra Singh and Salakhutdinov, Ruslan and Gupta, Abhinav and Gupta, Saurabh},
  booktitle = {CVPR},
  year      = {2020}
}

@inproceedings{armeni20193dsg,
  title     = {{3D} Scene Graph: A Structure for Unified Semantics, {3D} Space, and Camera},
  author    = {Armeni, Iro and others},
  booktitle = {ICCV},
  year      = {2019}
}

@inproceedings{hughes2022hydra,
  title     = {Hydra: A Real-time Spatial Perception System for {3D} Scene Graph Construction and Optimization},
  author    = {Hughes, Nathan and Chang, Yun and Carlone, Luca},
  booktitle = {RSS},
  year      = {2022}
}

@article{oquab2023dinov2,
  title   = {{DINOv2}: Learning Robust Visual Features without Supervision},
  author  = {Oquab, Maxime and others},
  journal = {Transactions on Machine Learning Research},
  year    = {2024}
}

@inproceedings{zhai2023siglip,
  title     = {Sigmoid Loss for Language Image Pre-Training},
  author    = {Zhai, Xiaohua and Mustafa, Basil and Kolesnikov, Alexander and Beyer, Lucas},
  booktitle = {ICCV},
  year      = {2023}
}

@article{qwen3,
  title   = {{Qwen3} Technical Report},
  author  = {Yang, An and others},
  journal = {arXiv preprint arXiv:2505.09388},
  year    = {2025}
}

@inproceedings{hu2022lora,
  title     = {{LoRA}: Low-Rank Adaptation of Large Language Models},
  author    = {Hu, Edward J. and others},
  booktitle = {ICLR},
  year      = {2022}
}

@inproceedings{zhou2019continuity,
  title     = {On the Continuity of Rotation Representations in Neural Networks},
  author    = {Zhou, Yi and Barnes, Connelly and Lu, Jingwan and Yang, Jimei and Li, Hao},
  booktitle = {CVPR},
  year      = {2019}
}

@article{bai2025qwen25vl,
  title   = {{Qwen2.5-VL} Technical Report},
  author  = {Bai, Shuai and others},
  journal = {arXiv preprint arXiv:2502.13923},
  year    = {2025}
}

@article{zhang2024llava,
  title={Llava-video: Video instruction tuning with synthetic data},
  author={Zhang, Yuanhan and Wu, Jinming and Li, Wei and Li, Bo and Ma, Zejun and Liu, Ziwei and Li, Chunyuan},
  journal={arXiv preprint arXiv:2410.02713},
  year={2024}
}

@inproceedings{liu2023llava,
  title     = {Visual Instruction Tuning},
  author    = {Liu, Haotian and Li, Chunyuan and Wu, Qingyang and Lee, Yong Jae},
  booktitle = {NeurIPS},
  year      = {2023}
}

@article{li2026ovo,
  title   = {{OVO-S-Bench}: A Hierarchical Benchmark for Streaming Spatial Intelligence in Multimodal {LLMs}},
  author  = {Li, Yifei and others},
  journal = {arXiv preprint arXiv:2606.03890},
  year    = {2026}
}

@article{lin2025ostbench,
  title   = {{OST-Bench}: Evaluating the Capabilities of {MLLMs} in Online Spatio-temporal Scene Understanding},
  author  = {Lin, Jingli and others},
  journal = {arXiv preprint arXiv:2507.07984},
  year    = {2025}
}

@inproceedings{li2025sti,
  title={Sti-bench: Are mllms ready for precise spatial-temporal world understanding?},
  author={Li, Yun and Zhang, Yiming and Lin, Tao and Liu, XiangRui and Cai, Wenxiao and Liu, Zheng and Zhao, Bo},
  booktitle={2025 IEEE/CVF International Conference on Computer Vision (ICCV)},
  pages={5622--5632},
  year={2025},
  organization={IEEE}
}

@inproceedings{yang2026mmsibench,
  title     = {{MMSI-Bench}: A Benchmark for Multi-Image Spatial Intelligence},
  author    = {Yang, Sihan and others},
  booktitle = {ICLR},
  year      = {2026}
}

@inproceedings{zhang2025spar,
  title     = {From Flatland to Space: Teaching Vision-Language Models to Perceive and Reason in {3D}},
  author    = {Zhang, Jiahui and others},
  booktitle = {NeurIPS Datasets and Benchmarks Track},
  year      = {2025}
}

@inproceedings{chen2026ttt3r,
  title={Ttt3r: 3d reconstruction as test-time training},
  author={Chen, Xingyu and Chen, Yue and Xiu, Yuliang and Geiger, Andreas and Chen, Anpei},
  booktitle={International Conference on Learning Representations},
  volume={2026},
  pages={50694--50718},
  year={2026}
}

@article{lin2025mmsi,
  title={Mmsi-video-bench: A holistic benchmark for video-based spatial intelligence},
  author={Lin, Jingli and Xu, Runsen and Zhu, Shaohao and Yang, Sihan and Cao, Peizhou and Ran, Yunlong and Hu, Miao and Zhu, Chenming and Xie, Yiman and Long, Yilin and others},
  journal={arXiv preprint arXiv:2512.10863},
  year={2025}
}

@misc{openai2026gpt6luna,
    author       = {{OpenAI}},
    title        = {Introducing {GPT-6 Sol and Luna}},
    year         = {2026},
    howpublished = {\url{https://openai.com/index/introducing-gpt-6-sol-and-luna/}}
  }

@misc{anthropic2026opus55,
    author       = {{Anthropic}},
    title        = {Introducing {Claude Opus 5.5}},
    year         = {2026},
    howpublished = {\url{https://www.anthropic.com/claude-opus-5-5}}
  }

@misc{anthropic2026sonnet55,
    author       = {{Anthropic}},
    title        = {Introducing {Claude Sonnet 5.5}},
    year         = {2026},
    howpublished = {\url{https://www.anthropic.com/claude-sonnet-5-5}}
  }

@misc{anthropic2026haiku55,
    author       = {{Anthropic}},
    title        = {Introducing {Claude Haiku 5.5}},
    year         = {2026},
    howpublished = {\url{https://www.anthropic.com/claude-haiku-5-5}}
  }

@article{openai2024gpt4o,
    author  = {{OpenAI}},
    title   = {{GPT-4o} System Card},
    journal = {arXiv preprint arXiv:2410.21276},
    year    = {2024}
  }

@misc{google2026gemini35flashlite,
    author       = {{Google}},
    title        = {{Gemini 3.5 Flash-Lite}},
    year         = {2026},
    howpublished = {\url{https://ai.google.dev/gemini-api/docs/models/gemini-3.5-flash-lite}},
    note         = {Accessed: 2026-11-16}
  }

@misc{google2026gemini31flashlite,
    author       = {{Google}},
    title        = {{Gemini 3.1 Flash-Lite}},
    year         = {2026},
    howpublished = {\url{https://ai.google.dev/gemini-api/docs/models/gemini-3.1-flash-lite}},
    note         = {Accessed: 2026-11-16}
  }

@article{li2024llava,
  title={Llava-onevision: Easy visual task transfer},
  author={Li, Bo and Zhang, Yuanhan and Guo, Dong and Zhang, Renrui and Li, Feng and Zhang, Hao and Zhang, Kaichen and Zhang, Peiyuan and Li, Yanwei and Liu, Ziwei and others},
  journal={arXiv preprint arXiv:2408.03326},
  year={2024}
}

@article{gu2024infinitymm,
    title   = {{Infinity-MM}: Scaling Multimodal Performance
               with Large-Scale and High-Quality Instruction Data},
    author  = {Gu, Shuhao and Zhang, Jialing and Zhou, Siyuan and others},
    journal = {arXiv preprint arXiv:2410.18558},
    year    = {2024}
  }
